%% file: main.tex
\documentclass[10pt,twocolumn,letterpaper]{article}

\usepackage[pagenumbers]{cvpr} 

\usepackage{graphicx}
\usepackage{amsmath}
\usepackage{amssymb}
\usepackage{booktabs}
\usepackage{xcolor}
\usepackage{lipsum}
\usepackage{ulem}
\usepackage{makecell}
\usepackage{booktabs,multirow,adjustbox,diagbox,threeparttable}
\usepackage[pagebackref,breaklinks,colorlinks]{hyperref}
\usepackage{enumitem}

\usepackage[pagebackref,breaklinks,colorlinks]{hyperref}

\usepackage[capitalize]{cleveref}
\crefname{section}{Sec.}{Secs.}
\Crefname{section}{Section}{Sections}
\Crefname{table}{Table}{Tables}
\crefname{table}{Tab.}{Tabs.}

\def\paperID{*****} 
\def\confName{CVPR}
\def\confYear{2024}

\input{prefix}
\begin{document}




\title{Gaussian Shell Maps for Efficient 3D Human Generation}

\author{Rameen Abdal\textsuperscript{*1} \quad Wang Yifan\textsuperscript{*1} \quad Zifan Shi\textsuperscript{*†1,2}  \quad  Yinghao Xu\textsuperscript{1} \quad Ryan Po\textsuperscript{1}  \quad Zhengfei Kuang\textsuperscript{1} 
\\ \\
 \quad Qifeng Chen\textsuperscript{2} \quad Dit-Yan Yeung\textsuperscript{2} \quad Gordon Wetzstein\textsuperscript{1} \\
\\
\textsuperscript{1}Stanford University \quad  \textsuperscript{2}HKUST 
}

\twocolumn[{%
\renewcommand\twocolumn[1][]{#1}%
\vspace{-2.1cm}
\maketitle
\thispagestyle{empty}
\begin{center}
     \includegraphics[width=\linewidth]{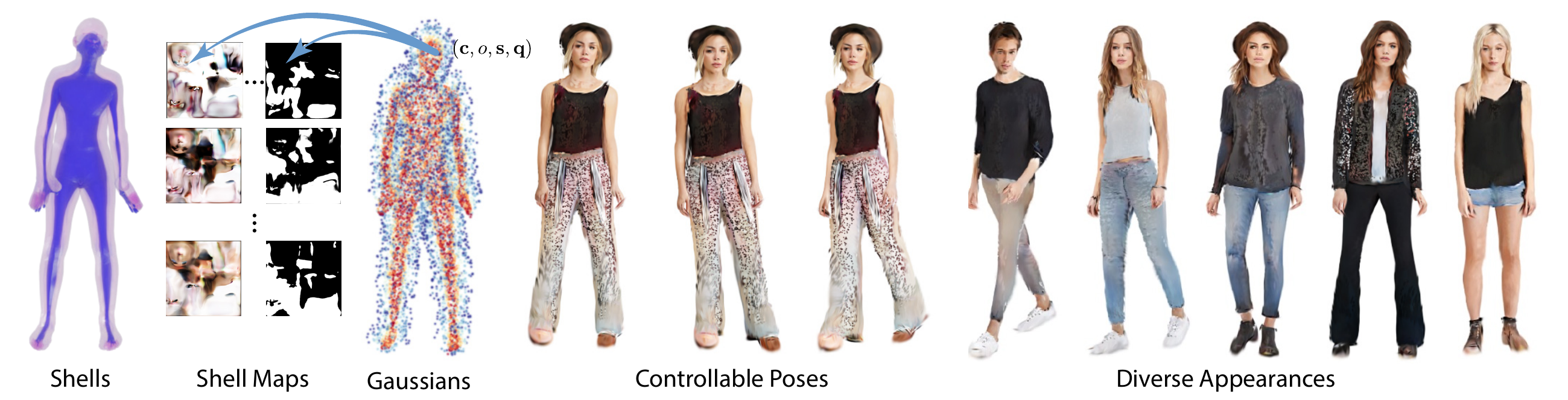}
     \captionof{figure}{\textbf{Gaussian Shell Maps.} Gaussian Shell Maps is an efficient framework for 3D human generation connecting 3D Gaussians with CNN-based generators. 3D Gaussians are anchored to ``shells" derived from the SMPL template~\cite{loper2015smpl} (only two shells are visualized for clarity), and the appearance is modeled in texture space. Trained only on 2D images, we show that our method can generate diverse articulable humans in real-time with state-of-the-art quality directly in high resolution without the need for upsampling and hence avoiding aliasing artifacts.}
    \label{fig:teaser}
\end{center}
}]

\begin{abstract}
Efficient generation of 3D digital humans is important in several industries, including virtual reality, social media, and cinematic production. 3D generative adversarial networks (GANs) have demonstrated state-of-the-art (SOTA) quality and diversity for generated assets. Current 3D GAN architectures, however, typically rely on volume representations, which are slow to render, thereby hampering the GAN training and requiring multi-view-inconsistent 2D upsamplers. Here, we introduce Gaussian Shell Maps (GSMs) as a framework that connects SOTA generator network architectures with emerging 3D Gaussian rendering primitives using an articulable multi shell--based scaffold. In this setting, a CNN generates a 3D texture stack with features that are mapped to the shells. The latter represent inflated and deflated versions of a template surface of a digital human in a canonical body pose. Instead of rasterizing the shells directly, we sample 3D Gaussians on the shells whose attributes are encoded in the texture features. These Gaussians are efficiently and differentiably rendered. The ability to articulate the shells is important during GAN training and, at inference time, to deform a body into arbitrary user-defined poses. Our efficient rendering scheme bypasses the need for view-inconsistent upsamplers and achieves high-quality multi-view consistent renderings at a native resolution of $512 \times 512$ pixels. We demonstrate that GSMs successfully generate 3D humans when trained on single-view datasets, including SHHQ and DeepFashion. \\
Project Page:  \href{https://rameenabdal.github.io/GaussianShellMaps/}{rameenabdal.github.io/GaussianShellMaps} 
\end{abstract}


\input{tex/introduction}

\input{tex/related}

\input{tex/background.tex}
\input{tex/method_new}

\input{tex/experiment}

\input{tex/conclusion}

\vspace{0.1cm}
\paragraph{Acknowledgements.} We thank Thabo Beeler and Guandao Yang for fruitful discussions and Alexander Bergman for help with baseline comparisons. We also thank the authors of GetAvartar for providing additional evaluation results. This work was in part supported by Google, Samsung, Stanford HAI, and the Swiss Postdoc Mobility Fund.


\input{tex/supplement}

{\small
\bibliographystyle{ieeenat_fullname}
\bibliography{egbib}
}

\end{document}

%% file: prefix.tex
\usepackage[font=small,skip=2pt]{caption}
\usepackage{makecell}
\usepackage{paralist}
\usepackage{color, colortbl}
\definecolor{LightGray}{gray}{0.9}

\newcommand{\supp}{\textit{Supplementary Materials}\xspace}

\newcommand{\moniker}{GSM\xspace}

\usepackage{multirow}
\usepackage{array}

\newcommand{\pose}{\boldsymbol{\theta}}
\newcommand{\shape}{\boldsymbol{\beta}}

\newcommand{\gPos}{\mathbf{x}}
\newcommand{\gOpc}{o}
\newcommand{\gCol}{\mathbf{c}}
\newcommand{\gScl}{\mathbf{s}}
\newcommand{\gRot}{\mathbf{q}}
\newcommand{\quat}{\mathbf{p}}
\newcommand{\Quat}{\mathbf{P}}
\newcommand{\pos}{\mathbf{x}'}
\newcommand{\col}{\gCol'}
\newcommand{\opc}{\gOpc'}

\newcommand{\template}{\mathbf{M}}

\renewcommand{\polygon}{\mathbf{T}}

\newcommand{\R}{\mathbb{R}}

\DeclareMathOperator{\smpl}{SMPL}

\DeclareMathOperator{\rot2quat}{rot2quat}

\newcommand{\D}{\mathcal{D}}

\newcommand{\jointregressor}{J}
\newcommand{\texture}{\mathsf{T}}
\newcommand{\mask}{\mathsf{M}}

\newcommand{\loss}[1]{\mathcal{L}_{\textrm{#1}}}
\newcommand{\tr}{\intercal}

\usepackage{xpunctuate}
\providecommand{\eg}[0]{e.g\xperiod}
\providecommand{\ie}[0]{i.e\xperiod}

\providecommand{\wrt}[0]{w.r.t\xperiod}

\usepackage{array}
\newcolumntype{L}[1]{>{\raggedright\let\newline\\\arraybackslash\hspace{0pt}}m{#1}}
\newcolumntype{C}[1]{>{\centering\let\newline\\\arraybackslash\hspace{0pt}}m{#1}}
\newcolumntype{R}[1]{>{\raggedleft\let\newline\\\arraybackslash\hspace{0pt}}m{#1}}

%% file: tex/introduction.tex
\vspace{-0.1cm}
\section{Introduction}
\label{sec:intro}
\footnote[0]{\textsuperscript{*} Equal Contribution}
\footnote[0]{\textsuperscript{†} Work done as a visiting student researcher at Stanford University}
The ability to generate articulable three-dimensional digital humans augments traditional asset creation and animation workflows, which are laborious and costly. Such generative artificial intelligence--fueled workflows are crucial in several applications, including communication, cinematic production, and interactive gaming, among others.

3D Generative Adversarial Networks (GANs) have emerged as the state-of-the-art (SOTA) platform in this domain, enabling the generation of diverse 3D assets at interactive framerates~\cite{noguchi2022unsupervised,bergman2022generative,hong2023evad,zhang2023getavatar,dong2023ag3d,cao2023dreamavatar,LayeredSurfaceVolumes,zhang2023getavatar}.
Most existing 3D GANs build on variants of volumetric scene representations combined with neural volume rendering~\cite{tewari2022advances}. However, volume rendering is relatively slow and compromises the training of a GAN, which requires tens of millions of forward rendering passes to converge~\cite{EG3D}. Mesh--based representations building on fast differentiable rasterization have been proposed to overcome this limitation~\cite{grigorev2021stylepeople,LayeredSurfaceVolumes}, but these are not expressive enough to realistically model features like hair, clothing, or accessories, which deviate significantly from the template mesh. These limitations, which are largely imposed by a tradeoff between efficient or expressive scene representations, have been constraining the quality and resolution of existing 3D GANs. While partially compensated for by using 2D convolutional neural network (CNN)--based upsamplers~\cite{StyleNeRF,EG3D}, upsampling leads to multi-view inconsistency in the form of aliasing.

Very recently, 3D Gaussians have been introduced as a promising neural scene representation offering fast rendering speed and high expressivity~\cite{kerbl3Dgaussians}. While 3D Gaussians have been explored in the context of single-scene overfitting, their full potential in generative settings has yet to be unlocked. This is challenging because it is not obvious how to combine SOTA CNN-based generators~\cite{Karras2019stylegan2,stylegan3} with 3D Gaussian primitives that inherently do not exist on a regular Cartesian grid and that may vary in numbers.

We introduce Gaussian Shell Maps (\moniker{}s), a 3D GAN framework that intuitively connects CNN generators with 3D Gaussians used as efficient rendering primitives.
Inspired by the traditional computer graphics work on shell maps~\cite{10.1145/1186822.1073239}, \moniker{}s use the CNN generator to produce texture maps for a set of ``shell'' meshes appropriately inflated and deflated from the popular SMPL mesh template for human bodies~\cite{loper2015smpl}.
The textures on the individual shells directly encode the properties of 3D Gaussians, which are sampled on the shell surfaces at fixed locations. The generated images are rendered using highly efficient Gaussian splatting, and articulation of these Gaussians can be naturally enabled through deforming the scaffolding shells with the SMPL model. Since 3D Gaussians have spatial extent, they represent details on, in between, and outside the discrete shells. \moniker{}s are trained exclusively on datasets containing single-view images of human bodies, such as SHHQ~\cite{fu2022stylehuman}.

Our experiments demonstrate that \moniker{} can generate highly diverse appearances, including loose clothing and accessories, at high resolution, without an upsampler, at a state-of-the-art rendering speed of 125~FPS (or 35FPS including generation).
Among various architecture design choices, multiple shells with fixed relative locations of the 3D Gaussian achieve the best results in our experiments.

Specifically, our contributions include
\begin{compactenum}
\item We propose a novel 3D GAN framework combining a CNN-based generator and 3D Gaussian rendering primitives using shell maps.
\item We demonstrate the fastest 3D GAN architecture to date, achieving real-time rendering of $512^2$~px without convolutional upsamplers, with image quality and diversity matching the state of the art on challenging single-view datasets of human bodies.
\end{compactenum}

\begin{figure*}[t!]
\centering
\noindent\includegraphics[width=\linewidth]{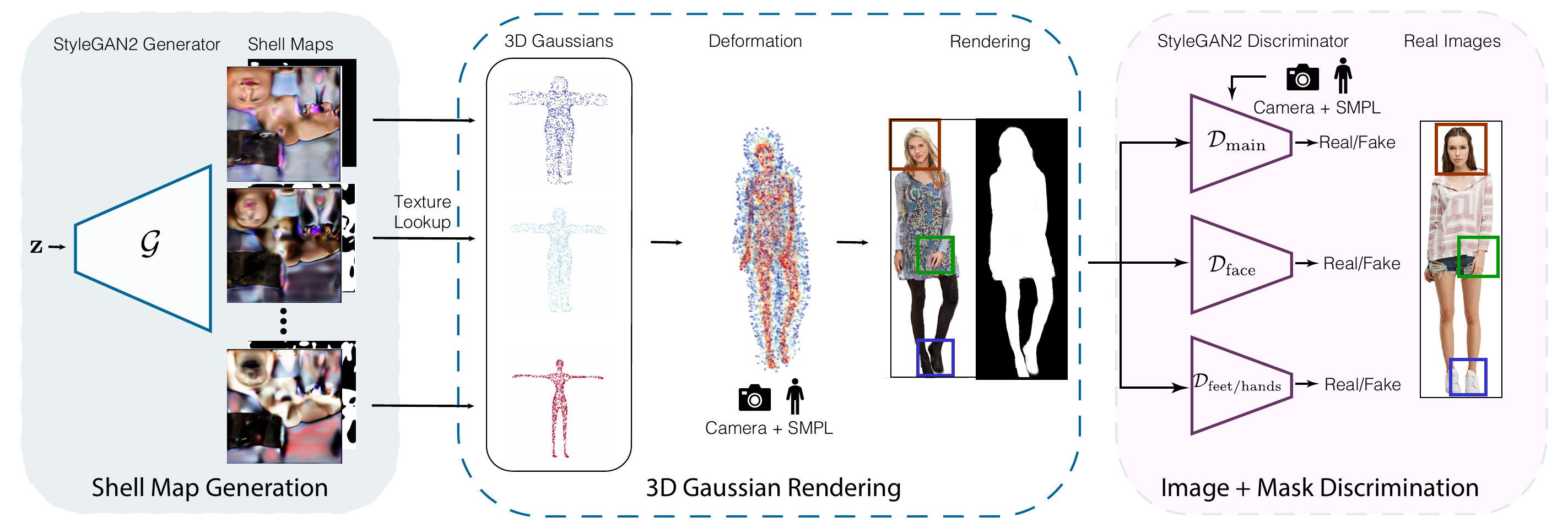}
\caption[Method Overview]{\textbf{Method Overview.} We propose an expressive yet highly efficient representation, Gaussian Shell Map (GSM), for 3D human generation. Combining the idea of 3D Gaussians and Shell Maps (\cref{sec:shell_map}), we sample 3D Gaussians on ``shells'', which are mesh layers offsetted from the SMPL template, forming a shell volume to model complex and diverse geometry and appearance; the Gaussian parameters are learned in the texture space, allowing us to leverage existing CNN-based generative architecture (\cref{sec:representation}). Articulation is straightforward by interpolating the deformation of the shell (\cref{sec:deformation}). The generation is supervised by single-view 2D images using several discriminator critics, including part-specific face, hands, and feet discriminators (\cref{sec:gan}).}
\label{fig:pipeline}
\end{figure*}

%% file: tex/related.tex
\section{Related Work } \label{sec:related}

\paragraph{3D-Aware Generative Models.}
GANs and diffusion models are two very powerful generative models emerging as a result of efficient architectures and high-quality datasets~\cite{StyleGAN,DALLE-2}. With the quality of image generation reaching photo-realism, the community started leveraging these SOTA generative models for 3D generation by combining neural rendering methods~\cite{tewari2020state,tewari2022advances} to produce high-quality multi-view consistent 3D objects from image collections. Several 3D GAN architectures have explored implicit or explicit neural volume representations for modeling 3D objects, including~\cite{GRAF, piGAN, CAMPARI, Giraffe, EG3D, StyleNeRF, VolumeGAN, StyleSDF, ide-3d,skorokhodov20233d,xu2022discoscene,an2023panohead,DNRGARD23}. 3D diffusion models, on the other hand, typically use the priors encoded by pre-trained text-to-image 2D diffusion models, e.g.~\cite{Saharia2022PhotorealisticTD, Balaji2022eDiffITD, Rombach2021HighResolutionIS,po2023compositional,Kolotouros2023DreamHumanA3, Bergman2023Articulated3H,Chan2023GenerativeNV,Lin2022Magic3DHT, Chen2023Fantasia3DDG, Tsalicoglou2023TextMeshGO, cao2023dreamavatar,poole2022dreamfusion, Liu2023Zero1to3ZO} (see this survey for more details~\cite{po2023state}). 
Due to the lack of high-quality, large-scale multi-view datasets curated for specific categories like humans, the choice of a suitable generative model becomes critical. 
Using diffusion models to generate high-resolution multi-view consistent 3D objects is still an unsolved problem~\cite{po2023state}.  3D GANs, on the other hand, exhibit better quality and multi-view consistency at higher resolutions that do not assume multi-view data~\cite{LayeredSurfaceVolumes,dong2023ag3d,EG3D,ide-3d}. This motivates our choice of building an efficient representation using a 3D GAN framework.
\paragraph{Generative Articulated 3D Digital Humans.}
3D-GAN frameworks have been proposed to generate the appearance, geometry, and identity-preserving novel views of digital humans~\cite{bergman2022generative, LayeredSurfaceVolumes,hong2023evad, Avatargen2023,zhang2023getavatar,dong2023ag3d,chen2023veri3d,yang20233dhumangan}. Most of these GANs are trained on single-view image collections~\cite{liu2016deepfasion,li2021aist,fu2022stylegan}. A popular approach is to use a neural radiance representation~\cite{bergman2022generative,noguchi2022unsupervised,jiang2023humangen,Avatargen2023, hong2023evad} where a canonically posed human can be articulated via deformation. Other approaches are based on meshes, where a template can be fixed or learned during the training~\cite{grigorev2021stylepeople,yang20233dhumangan,sun2023next3d,aneja2022clipface,gao2022get3d}.
Related to this line of work, a concurrent paper, LSV-GAN~\cite{LayeredSurfaceVolumes}, offsets the SMPL template meshes into layered surfaces and composites the per-layer rasterization result to form the final rendering. While it provides a faster alternative to the volume rendering-based approaches, it can only accommodate a small offset from the SMPL mesh, which hampers diversity. Our \moniker{} method differs from LSV-GAN as we employ 3D Gaussians~\cite{kerbl3Dgaussians} as primitives on the layered surfaces, which, by having a learnable spatial span, allows for larger deviation from the template mesh and can thus generate more intricate details.
\paragraph{Point-Based Rendering.}
Earlier point-based methods efficiently render point clouds and rasterize them by fixing the size~\cite{gross2011point,grossman1998point,sainz2004point}. While efficient in terms of speed and parallel rasterization on graphics processing units~\cite{schutz2022software,laine2011high}, they are not differentiable~\cite{ren2002object, kerbl3Dgaussians}. To combine these methods with the neural networks and perform view synthesis, recent works have developed differentiable point-based rendering techniques~\cite{yifan2019differentiable,wiles2020synsin,ruckert2022adop,aliev2020neural,xu2022point,zhang2022differentiable, zheng2023pointavatar}. More recently, 3D Gaussian-based point splatting gained traction due to the flexibility of anisotropic covariance and density control with efficient depth sorting~\cite{kerbl3Dgaussians}. This allows 3D Gaussian splats to handle complex scenes composed of high and low-frequency features. Relevant to human bodies, 3D Gaussians have also
been used in pose estimation and tracking~\cite{GaussiGAN,stoll2011fast,rhodin2015versatile}. While the Gaussian primitives have been used for efficient scene reconstruction and novel view synthesis, it is not trivial to deploy Gaussians in a generative setup. To the best of our knowledge, our method is the first to propose a combination of 3D Gaussians and 3D GANs.

%% file: tex/background.tex
\begin{figure*}[t!]
\centering
\noindent\includegraphics[width=\linewidth, trim={0 8ex 0 4ex}, clip]{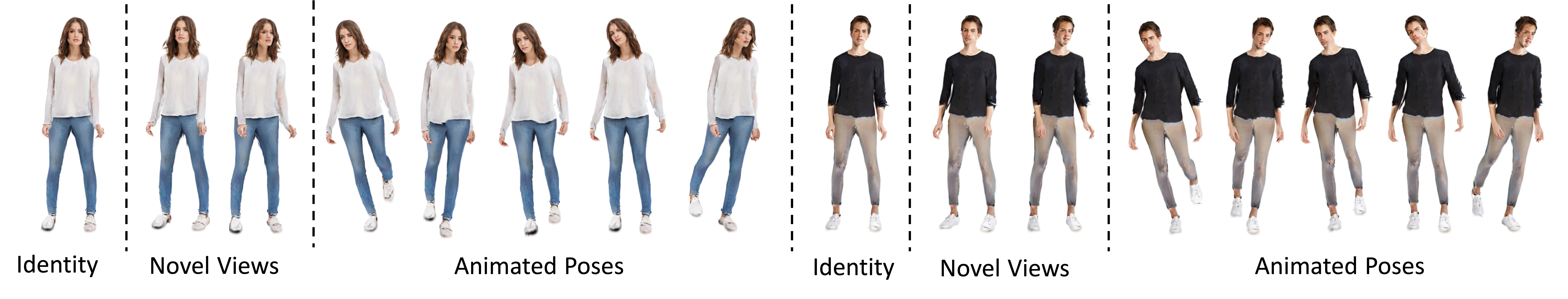}
\caption{\textbf{Novel Views and Animation.} We can render a generated identity in novel views and articulate it for animations.}
\label{fig:animation}
\end{figure*}

\section{Background}\label{sec:background}

\paragraph{3D Gaussians.}
\label{sec:gaussian_overview}
These point-based primitives can be differentiably and efficiently rendered using EWA (elliptical weighted average) volume splatting~\cite{zwicker2001ewa}.
3D Gaussians have recently demonstrated outstanding expressivity for 3D scene reconstruction~\cite{kerbl3Dgaussians}, in which the Gaussian parameters, position $\gPos \in \R^3$, opacity $\gOpc\in \R$, color $\gCol\in \R^{sh}$ ($sh$ representing the spherical harmonic coefficients), scaling $\gScl\in \R^3$ , and rotation $\gRot\in \R^4$ parameterized as quaternions are jointly optimized to minimize the photometric errors of the rendered images in a set of known camera views.
The optimization is accompanied by adaptive control of the density, where the points are added or removed based on the density, size, and gradient of the Gaussians.

Specifically, each Gaussian is defined as
\begin{equation}
G(\pos;\gPos, \Sigma)=\exp^{-\frac{1}{2}\left( \pos-\gPos \right)^\tr \Sigma^{-1}\left( \pos-\gPos \right)},\label{eq:gaussian_def}
\end{equation}
where $\Sigma=R S S^T R^T$ is the covariance matrix parameterized by the rotation and scaling matrices $R$ and $S$ given by the quaternions \(\gRot\) and scaling \(\gScl\).

The image formation is governed by classic point-based \(\alpha\)-blending of overlapping Gaussians ordered from closest to farthest~\cite{kopanas2021point}:
\begin{equation}
\mathbf{C} =\sum_{i \in \mathcal{N}} \col_i\opc_i \prod_{j=1}^{i-1}\left(1-\mathbf{\alpha_j}\right),\label{eq:alpha_blending}
\end{equation}
where color $\col_i$ and opacity $\opc_i$ is computed by 
\begin{equation}
\col_i = G(\pos;\gPos_i)\gCol_i\quad \textrm{and}\quad \opc_i = G(\pos;\gPos_i)\gOpc_i\label{eq:color_opacity}.
\end{equation}

Thanks to their ability to fit complex geometry and appearance, 3D Gaussians are gaining popularity for 3D scene reconstruction. 
However, deploying 3D Gaussians for generative tasks remains an unexplored topic and is challenging for several reasons. First, Gaussians are of ``Lagrangian'' nature -- their sparse number and learnable positions are challenging to combine with SOTA ``Eulerian'' (i.e., grid-based) CNN generators. Second, the parameters of Gaussians are highly correlated. The same radiance field can be equally well explained by many different configurations of Gaussians, varying their locations, sizes, scales, rotations, colors, and opacities. This ambiguity makes it challenging to generalize over a distribution of objects or scenes, which generative methods do. 

\paragraph{Shell Maps.}\label{sec:shell_map}
Our representation is related to Shell Maps~\cite{10.1145/1186822.1073239,DNRGARD23,sin2023nerfahedron}, a technique in computer graphics designed to model near-surface details.
Shell maps use 3D texture maps to store the fine-scale features in a shell-like volume close to a given base surface, typically represented as a triangle mesh.
In essence, they extend UV maps such that every point in the shell volume can be bijectively mapped to the 3D texture map for efficient modeling and rendering in texture space.
The shell volume is constructed by offsetting the base mesh along the normal direction while maintaining the topology and avoiding self-intersection.
The volume is discretized into tetrahedra that connect the vertices of the base and offset meshes.
A unique mapping between the shell space and the texture space can be established by identifying the tetrahedron and subsequently querying the barycentric coordinates inside it.
Formally, the mapping from \(\gPos\) in the shell volume to its position in the 3D texture map is defined as
\begin{equation}
\gPos_{t} = \phi\left(\polygon_{t}, B\left( \gPos, \polygon \right) \right),\label{eq:shell_map}
\end{equation}
where \(\polygon, \polygon_{t}\) refer to corresponding tetrahedra in the shell and texture space, \(B\left( \gPos, \polygon \right)\) is the barycentric coordinates of \(\gPos\) in \(\polygon\), and \(\phi\) denotes barycentric interpolation.

%% file: tex/method_new.tex
\begin{figure}
\centering
\noindent\includegraphics[width=\linewidth]{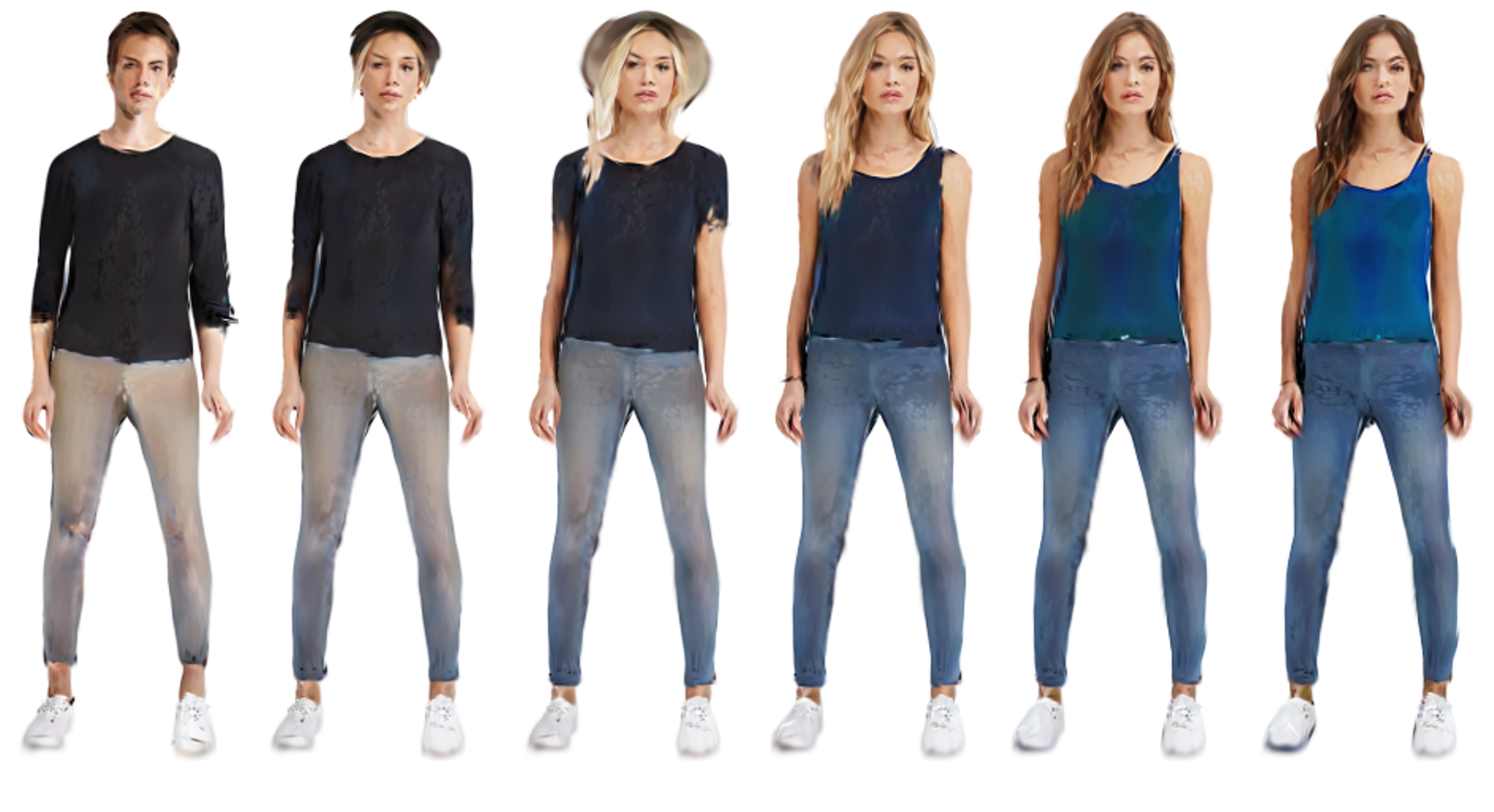}
\caption{\textbf{ Latent Code Interpolation.} Latent code interpolation of our model trained on DeepFashion Dataset.}
\label{fig:interpolation}
\end{figure}

\begin{figure*}
\centering

\noindent\includegraphics[width=\linewidth]{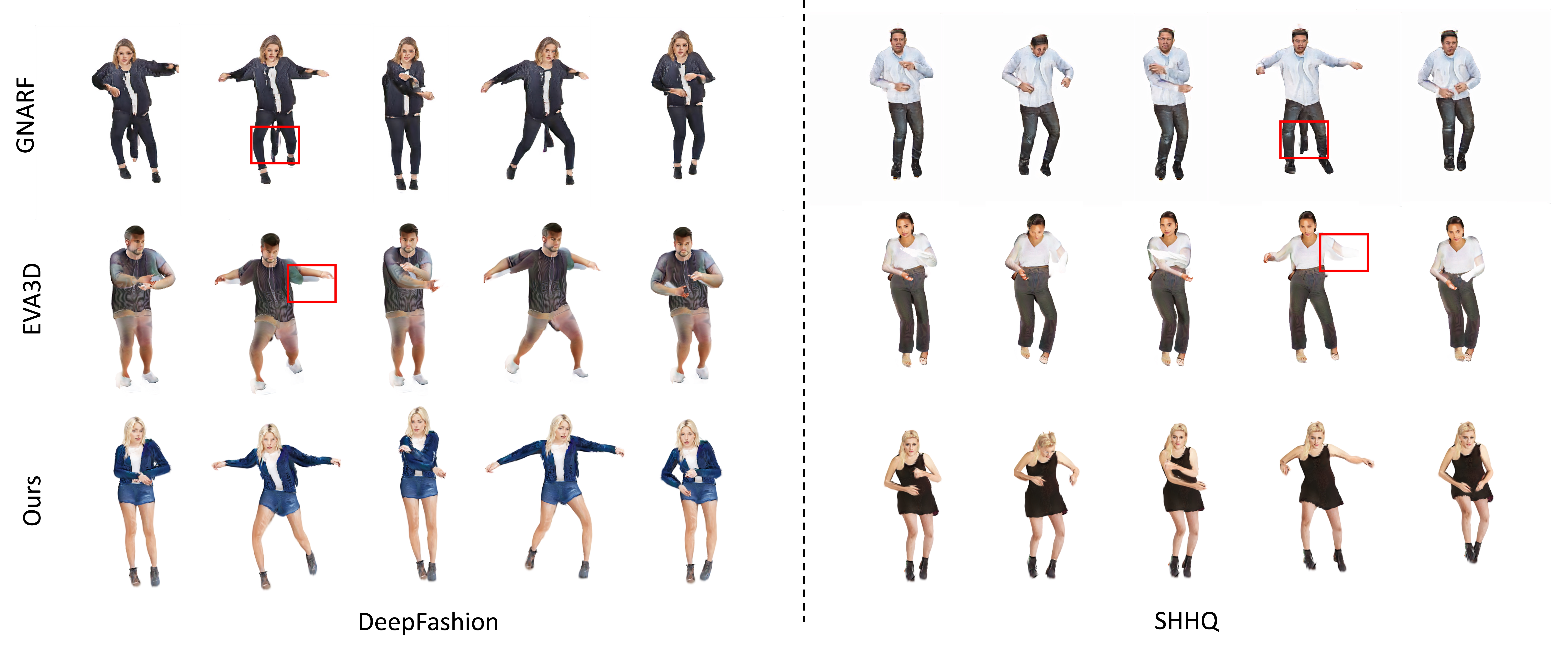}
\caption{\textbf{Qualitative Comparison.} We compare our results with GNARF and EVA3D  baselines on DeepFashion and SHHQ datasets. In each case, we show the deformed body poses of the identities generated by the methods. The competing methods exhibit artifacts marked in red. Notice that our approach generates high-quality textures, like facial details and more realistic deformations.}
\label{fig:comparison}
\end{figure*}

\section{Gaussian Shell Maps (GSM)}\label{sec:main_method}
\moniker{} is a framework that connects 3D Gaussians with SOTA CNN-based generator networks~\cite{EG3D} in a GAN setting.
The key idea is to anchor the 3D Gaussians at fixed locations on a set of ``shells'' derived from the SMPL~\cite{loper2015smpl} human body template mesh.
These shells span a volume to model surface details that deviate from the unclothed template mesh.
We learn the relevant Gaussian parameters in the texture space, allowing us to leverage established CNN-based generative backbones while seamlessly utilizing the parametric deformation model for articulation.
The overall pipeline is shown in~\Cref{fig:pipeline}.

\subsection{Representation}\label{sec:representation}
Our method utilizes the concept of shell maps to leverage the inherent planar structure of texture space for seamless integration with CNN-based generative architectures and, at the same time, encapsulate diverse and complex surface details without directly modifying the template mesh.
Specifically, the shell volume is defined by the boundary mesh layers, created by inflating and deflating the T-pose SMPL mesh using Laplacian Smoothing~\cite{sorkine2004laplacian} with the smoothing factor set to negative for inflation and positive for deflation.
We then represent this shell volume using \(N\) mesh layers, \ie ``shells'', by linearly spacing the vertices between the aforementioned boundary shells.
In parallel, we apply a similar discretization strategy to the 3D texture space, creating \(N\) 2D texture maps storing neural features that can be referenced for each shell using the UV mapping inherited from the SMPL template.

As shown in \Cref{fig:pipeline}, we use the shell maps to generate all Gaussian parameters except the positions, as those are sampled and anchored \wrt to the shells at every iteration (explained below in detail).
This results in a feature volume of \(\texture^{N\times H\times W\times 11}\), comprised of 3D color \(\texture^{\gCol}\), 1D opacity \(\texture^{\gOpc}\), 3D scaling \(\texture^{\gScl}\), and 4D rotation \(\texture^{\gRot}\) features.

We create Gaussians in our shell volume.
This is done by sampling a fixed number of Gaussians quasi-uniformly on the shells based on the triangle areas.
Once sampled, the Gaussians are anchored on the shells using barycentric coordinates so that every Gaussian center \(\gPos\) can be mapped to a point \(\gPos_t\) on the corresponding shell map using~\cref{eq:shell_map}, except that the tetrahedra are replaced with the triangle in which the Gaussian resides.
This anchoring is a crucial design choice, as it enables straightforward feature retrieval from the shell maps.
More importantly, it simplifies the learning by fixing the Gaussian positions and offloading geometry modeling to opacity \(\gOpc\) and sigma \(\Sigma\).

The spatial span of the Gaussians plays a significant role in reconstructing the features defined on the discrete shells into a continuous signal within the 3D space.
It enables every point---whether on the shells, between them, or outside them---to receive a valid opacity and color value by evaluating \cref{eq:color_opacity}.
This process allows us to model diverse appearances and body shapes, excelling mesh-based representation while at the same time maintaining efficient rendering, which is critical for GAN training.

Formally, Gaussian opacity, color, scale, and rotation can be interpolated from the shell maps \(\texture\)
\begin{equation}
    f = \texture^{f}\left(\gPos_t\right)\label{eq:gaussian_feature}, \textrm{ where } f=\left\lbrace\gOpc, \gCol, \gScl, \gRot\right\rbrace.
\end{equation}

\subsection{Deformation}
\label{sec:deformation}
The deformation step updates the Gaussians' locations and orientations based on the SMPL template mesh \(\template\), pose \(\pose\), and shape \(\shape\).
Note that, different from SMPL, our template mesh could be any of the shells.
Since each Gaussian is anchored on the shells using barycentric coordinates, we can query its new location and orientation simply from the associated vertices.
In particular, given the barycentric coordinates \(B\left(\gPos, \polygon\right)\) of a Gaussian inside triangle \(\polygon\), and the deformed position \(\polygon_{\textrm{new}}\) and the rotation quaternions \(\Quat\) on the vertices, we can obtain the new location and orientation of the Gaussian as
\begin{gather}
    \gPos_{\textrm{new}} =  \phi\left( \polygon_{\textrm{new}}, B\left( \gPos, \polygon \right) \right)\label{eq:deform_xyz}, \\
    \gRot_{\textrm{new}} = \dfrac{\hat{\quat}}{\left\Vert\hat{\quat} \right\Vert}\gRot,\textrm{where }\hat{\quat} = \phi\left( \Quat, B\left( \gPos, \polygon \right) \right). \label{eq:deform_quat}
\end{gather}
The deformed mesh is given by the SMPL deformation model \(\smpl\), and the quaternions \(\Quat\) are a result of the linear blend skinning (LBS) from regressed joints and skinning weights,
 \(\left\lbrace w_{j}\right\rbrace_{1}^{N_J}\), \ie,
\(\template_{\textrm{new}} = \smpl\left( \pose, \shape, \template \right)\) and \(\quat = \rot2quat\left( \sum_{j=1}^{N_J}w_{j}R_{j}\left( \pose, \jointregressor\left( \shape \right) \right)\right)\), where \(R_{j}\) is the rotation matrix of the \(j\)-th joint, and \(\jointregressor\) is the regressor function that maps the shape parameters to the joint locations.

\subsection{GAN training}
\label{sec:gan}

\paragraph{Generator.} Similar to prior work~\cite{EG3D}, we adopt StyleGAN2 without camera and pose conditioning~\cite{bergman2022generative,LayeredSurfaceVolumes} for the generator.
We use separate MLPs with different activations for $\gCol,\gOpc,\gRot$, and \(\gScl\): for $\gCol$, we use shifted sigmoid following the practice proposed in Mip-NeRF~\cite{mip-NeRF}; for $\gOpc$ we use sigmoid to constrain the range to \(\left(0,1\right)\); for $\gRot$ we normalize the raw MLP output to ensure they are quaternions (see~\cref{eq:deform_quat}); for $\gScl$ we use clamped exponential activation to limit the size of the Gaussians, which is critical for convergence based on our empirical study.

\paragraph{Discriminator.} Our discriminator closely follows that of \cite{bergman2022generative}. Since it does not have an upsampler, the input to the discriminator is the RGB image concatenated with the alpha channel (foreground mask), which is rendered using Gaussian rasterization with the Gaussian color set to 1 for fake samples and precomputed using off-the-shelf segmentation network~\cite{sandler2018mobilenetv2} for the real samples.
We refer to the discriminator using foreground mask ``Mask Discriminator''.
Including the alpha channel helps prevent the white background from bleeding into the appearance, which causes artifacts during articulation.

\paragraph{Face, Hand, and Feet Discriminator.} As the human body and clothes are diverse, the discriminator may choose to focus on these features and provide a weak signal to the facial, hand, and foot areas, which are crucial for visually appealing results. To avoid this problem, we adopt a dedicated Face Discriminator $\D_{\textrm{face}}$, Feet-and-Hands Discriminator $\D_{\textrm{feet/hands}}$ in addition to the main Discriminator $\D_{\textrm{main}}$.
All of these part-focused discriminators share the same base architecture as the main discriminator, except that we no longer use any conditioning since these cropped images do not contain distinct pose information.
The inputs are crops of the corresponding parts, whose spatial span is determined from the SMPL pose and camera parameters.
\vspace{-0.3cm}
\paragraph{Scaling Regularization.}
In our empirical study, we observe when unconstrained, the network tends to learn overly large or extremely small Gaussians early on during the training and rapidly leads to divergence or model collapse.
We experimented with multiple regularization strategies and different activation functions and found the following scaling regularization most effective:
\begin{gather}
    \loss{scale} = \mask\circ\|\texture^{\gScl} - s_{\textrm{ref}}\|^{2}\label{eq:scale_reg} \textrm{ with } s_{\textrm{ref}} = \frac{1}{P}\sum_{i=1}^{P} \ln\left( \delta_{i} \right),
\end{gather}
where \(\mask\) is the binary mask indicating UV-mapped region on the shell maps, and \(s_{\textrm{ref}}\) is a reference scale determined by the closest-neighbor distance \(\delta\) averaged among all the Gaussians.

%% file: tex/experiment.tex
\begin{table}
\center
\caption{\textbf{Quantitative Evaluation.} We compare our method with 3D-GAN baselines using DeepFashion and SHHQ datasets. We compute the FID score to evaluate the quality and diversity of the generated samples. Notice that our scores are comparable to state-of-the-art methods. To evaluate deformation consistency, we compute the PCK metric, where our approach consistently outperforms the baselines. INF. represents the inference speed measured in ms/img on an A6000 GPU at $512^2$ resolution. Note our method is the fastest across all competing methods. $\ast$ numbers are adopted from EVA3D~\cite{hong2023evad}; NA represents Not Available; --- represents Not Applicable, and  $+$ numbers are provided by the authors. }

\label{tab:main-results}
\vspace{-10pt}
\small
\begin{tabular}{lccccc}
\multicolumn{6}{c}{} \\
\toprule
\multirow{2}{*}{Model} & \multicolumn{2}{c}{Deep Fashion} & \multicolumn{2}{c}{SHHQ} & \multicolumn{1}{c}{Comp.} \\
\cmidrule(lr){2-3}
\cmidrule(lr){4-5}
\midrule
 &  FID $\downarrow$   &  PCK $\uparrow$  & FID $\downarrow$   &  PCK $\uparrow$  & INF. $\downarrow$\\
\midrule

EG3D$^*$   & 26.38 & --- & 32.96 & --- & 38\\
StyleSDF$^*$  & 92.40 & --- & 14.12 & --- & 32 \\ \midrule
ENARF$^*$  & 77.03 & 43.74 & 80.54 &  40.17 &  104  \\
GNARF   & 33.85 & 97.83 & 14.84 & \underline{98.96} &  72\\
EVA3D$^*$   & 15.91 & 87.50 & \bf{11.99} & 88.95 &  200\\
StylePeople   & 17.72 & \underline{98.31} & 14.67 & 98.58 &  \bf{28} \\

GetAvatar    & \hspace{0.2cm}$19.00^+$ & NA& NA &  NA &  44 \\

AG3D  & {\bf 10.93} & NA& NA &  NA  & 105 \\

Ours  & \underline{15.78} & \bf{99.48}& \underline{13.30} & \bf{99.27} & \bf{28} \\

\bottomrule
\end{tabular}
\vspace{-7pt}
\end{table}

\section{Experiment Settings}
\subsection{Datasets}\label{sec:dataset}
We evaluate our method using the two most common human datasets,  DeepFashion~\cite{liu2016deepfasion} and SHHQ~\cite{fu2022stylehuman}. DeepFashion and SHHQ do not provide SMPL parameters, so we use SMPLify-X~\cite{SMPL-X:2019} to obtain the SMPL parameters and camera poses.
\subsection{Training Details}\label{sec:implementation_details}
By default, we generate the shell maps at \(512\times512\) resolution for $8$ shells with a total of $100k$ Gaussians equally distributed across the shells. 
The offset between adjacent shells is \(0.08\).
The last fully-connected layer of the scaling prediction is adjusted to ensure the initial scaling is within a reasonable range, which is determined similar to \(s_\textrm{ref}\) in \cref{eq:scale_reg}.
The loss weights for all discriminators are set to 1, whereas the scaling regularization is set to \(0.1\).
$R1$ regularization is used for all the discriminators with a weight of $10$.
We apply progressive training starting at \(256\times 256\) rendering resolution for \(6000k\) training images and progressively grow the resolution to \(512\times 512\) for \(1000k\) images, then continue training at the fixed \(512\times 512\) resolution for \(3000k\) images.
We train our method on 8 A100 GPUs with a batch size of 32.
The learning rates are initialized at $0.002$ for both the generator and discriminator.

\subsection{Evaluation Details}
\paragraph{Baselines.}
We compare with the following volume rendering and rasterization-based methods: EG3D~\cite{EG3D}, StyleSDF~\cite{StyleSDF}, GNARF~\cite{bergman2022generative}, ENARF~\cite{noguchi2022unsupervised}, StylePeople~\cite{grigorev2021stylepeople},  EVA3D~\cite{hong2023evad}, GetAvatar~\cite{zhang2023getavatar}, and AG3D~\cite{dong2023ag3d}.
Among these methods, EG3D and StyleSDF do not support articulation; except StylePeople, all methods use neural field and volumetric rendering.
Regardless of representation, all these baselines use 2D convolution for post-rendering upsampling and/or postprocessing, causing flickering artifacts (see \supp).

\vspace{-0.2cm}
\paragraph{Metrics}
Following prior work, We use two metrics to evaluate the quality, diversity, and quality of animations: Fréchet Inception Distance (FID) and Percentage of Correct Keypoint (PCK). The former measures the visual quality and diversity of the generated samples (using $50k$ samples and the full dataset). The latter measures how close the generated image aligns with the pose control. We compute PCK on $5k$ samples using the implementation provided by GNARF~\cite{bergman2022generative}.

\begin{table}[t!]%
\caption{ 
\textbf{Ablation: Number of Shells.} Ablation on the number of shells performed on $128^2$ resolution trained for \(1600k\) images. }

\label{tab:ablation_shells}
\begin{minipage}{\columnwidth}
\begin{center}
\begin{tabular}{rccccc}
\toprule
$\#$ of Shells & 1 & 2 & 8 & 10 & 15\\ \midrule
FID $\downarrow$ & 40.3 & 25.31 & 18.70 & 21.57 & 21.10 \\\bottomrule
\end{tabular}
\end{center}
\end{minipage}
\end{table}%

\vspace{-0.2cm}

\section{Experiment Results}\label{sec:results}

\vspace{0.2cm}

\subsection{Qualitative}\label{sec:qualitative_results}
\Cref{fig:teaser} showcases some of our generated results. Notice that our model is able to generate diverse attributes, including loose clothing and accessories like hats.
Further, in \Cref{fig:animation}, we visualize the generated results in varying camera views and body poses.
To demonstrate that the latent space of the proposed model is smooth, we show visual results of latent code interpolation in \Cref{fig:interpolation} computed on the DeepFashion trained model.

\Cref{fig:comparison} shows the visual comparison of our method with the baselines on DeepFashion and SHHQ datasets, with more examples to be found in \supp.
Volumetric methods, such as GNARF, operate in a canonical pose and rely on accurate correspondence matching between the observed space and the canonical space, which is ill-defined; thus, these methods tend to produce artifacts for limbs, which undergo large deformation and occlusion. Our method is able to model the complex geometry and appearance of the human body, generating characters with loose clothing and accessories (see \Cref{fig:teaser}) and producing convincing results under various articulations.

\begin{figure}[t!]
\centering
\noindent\includegraphics[width =\linewidth]{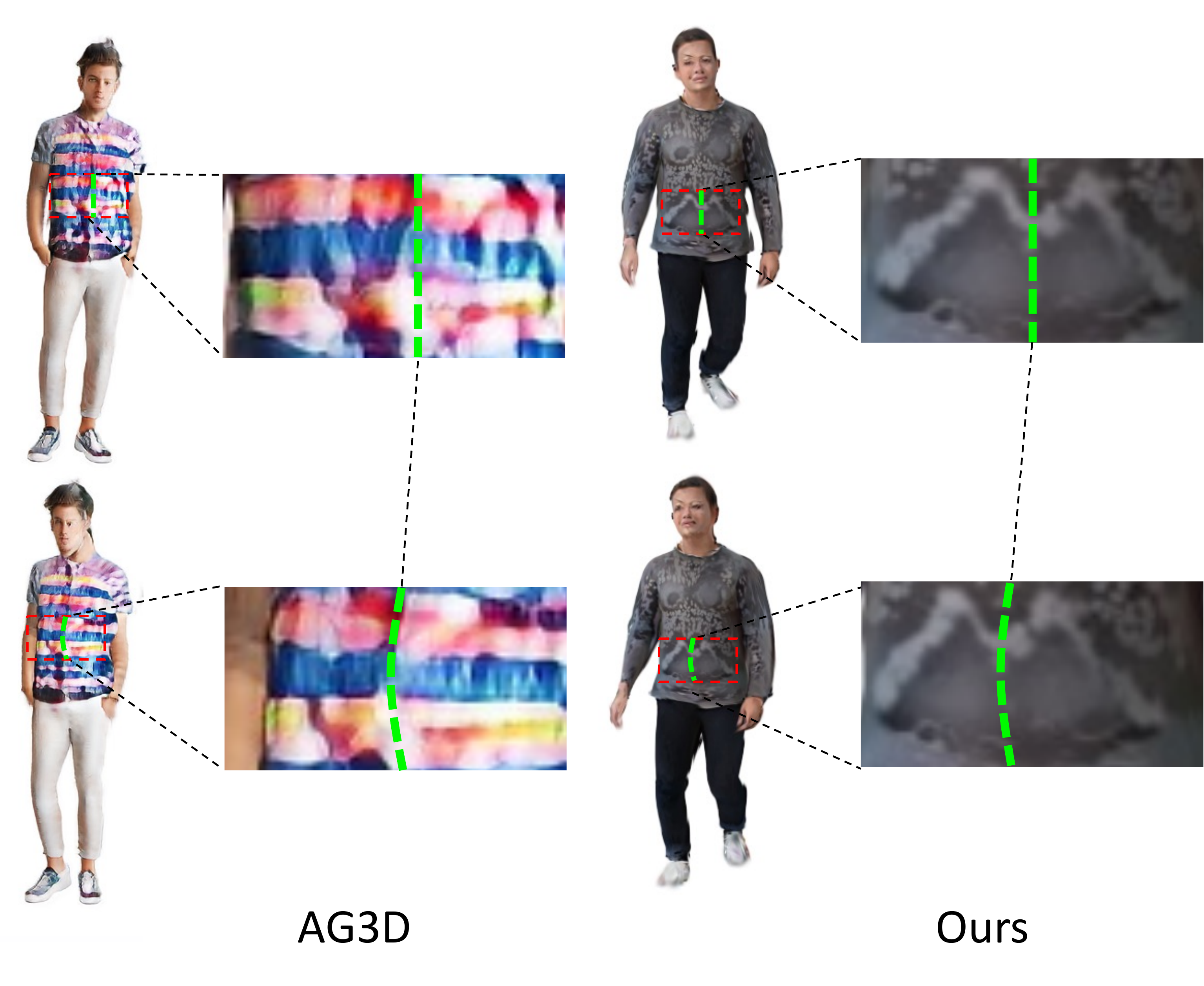}
\vspace{-20pt}
\caption{\textbf{Multi-view Consistency.} The figure shows a closeup of the garment in two different views. The green dashed line follows the body shape and indicates the same 3D position. In AG3D~\cite{dong2023ag3d}, the pattern significantly changes, while ours, utilizing texture maps, has built-in view consistency.}
\label{fig:ag3d_comparison}
\end{figure}
\vspace{0.2cm}
\subsection{Quantitative} In Table~\ref{tab:main-results}, we show the results of our method compared to the competing methods. 
Many of the methods have not released training or data processing code; we have collected these numbers with our best effort from the authors directly.

Our method performs on par with the SOTA methods in terms of visual quality and diversity measured by FID while ranking top in terms of inference speed. By utilizing the texture space, our model can control the articulation in a straightforward manner, contributing to consistent alignment with the pose control input, as reflected by the PCK score. Note that GetAvatar is trained on multiview data, and AG3D uses normal maps for supervision.

All other baseline methods adopt post-rendering convolution layers to upsample and add texture details, leading to severe aliasing artifacts not reflected in the quantitative evaluation.
We demonstrate this in~\Cref{fig:ag3d_comparison}, where we render a generated identity in two different views using our method and AG3D, which achieves the lowest FID on the DeepFashion dataset. Notice that with view change, the features produced by AG3D change significantly, leading to unnatural flickering in animation even though the FID is low.

\begin{figure}
  \centering
  \includegraphics[width=\linewidth]{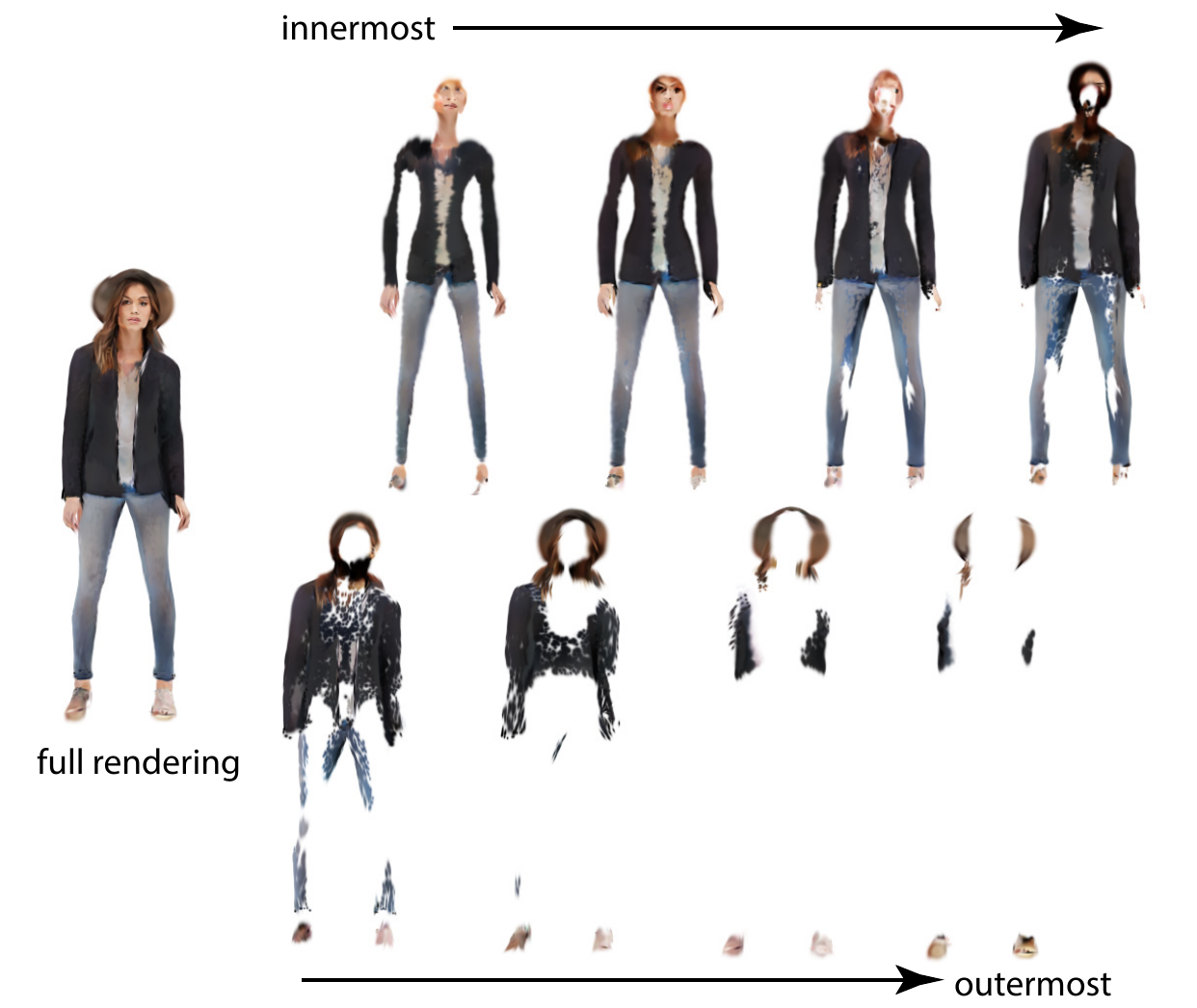}
  \caption{\textbf{Per-shell Visualization}. We visualize the per-shell contributions. The shells decompose the human body into different parts, where the inner shells capture the torso, and the outer shells capture the clothing and hair.}
  \label{fig:shell_visualization}
\end{figure}

\begin{figure}
\centering
\noindent\includegraphics[width=\linewidth]{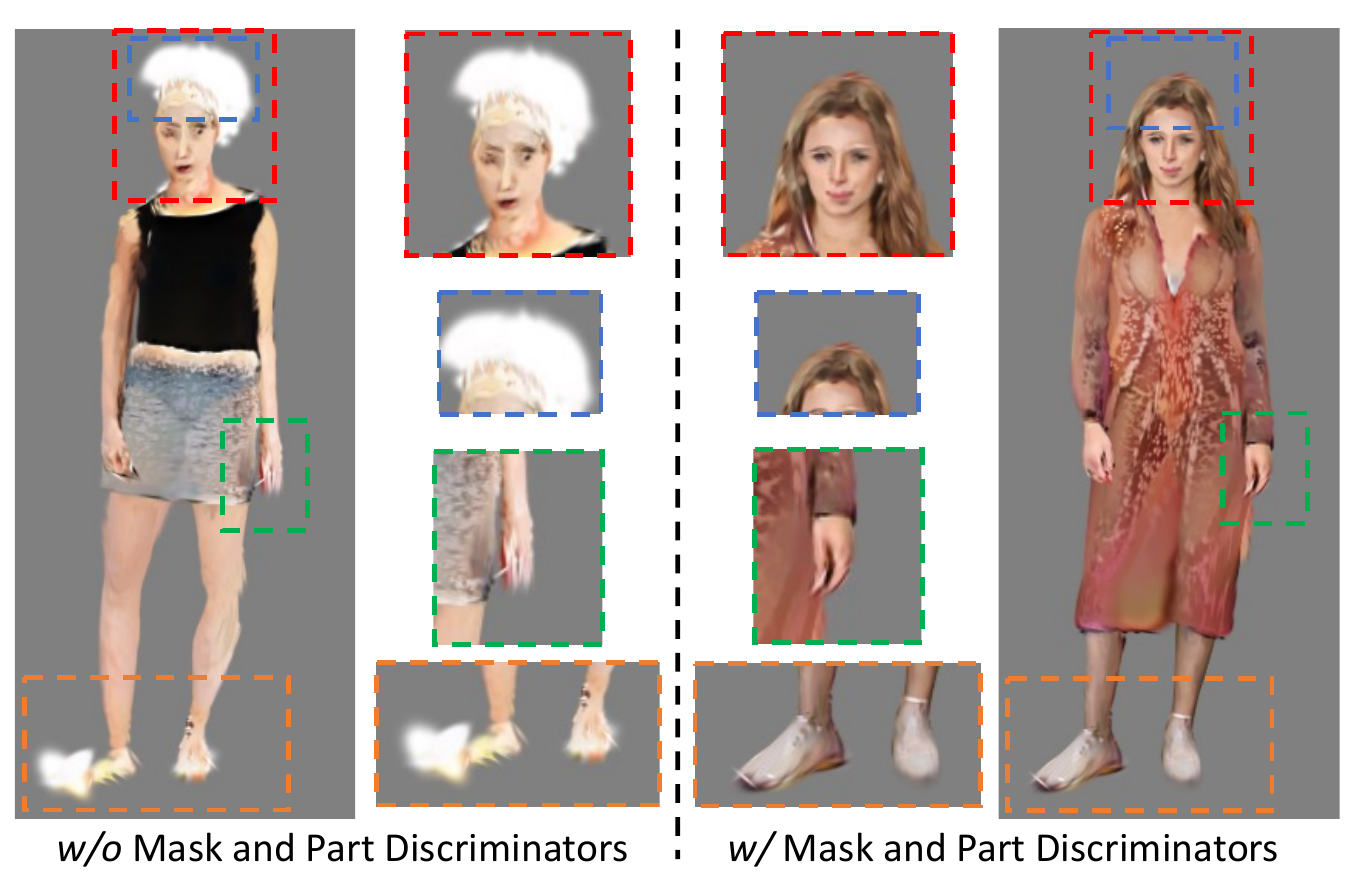}
\caption{\textbf{Mask + Part Discriminators.} The blue box highlights the improved silhouette after concatenating the alpha channels in the input to the discriminator. The other boxes showcase better details using the face, feet, and hands discriminator.}
\label{fig:ablation}
\end{figure}
\subsection{Ablation}\label{sec:ablation}
\vspace{0.2cm}
We conduct ablation studies on different configurations and components of our method to understand their effect on the final model's performance. To identify the trends, we test these configurations at $128^2$ resolution on the SHHQ dataset, trained for \(1600k\) images without progressive training. \(200k\) Gaussians are sampled for all the experiments.

\paragraph{Number of Shells.} The number of shells determines the granularity of variation we can model inside the shell volume, thus impacting the expressivity of our representation. 
We investigate this effect in \Cref{tab:ablation_shells}. 
With only one shell, although the Gaussians can vary in size and cover regions off the shell surface, they still lack the capacity to model diverse human body shapes and appearances.
Using two shells essentially defines the boundary for the shell volume. This variation significantly outperforms the single-shell variation, demonstrating the importance of shell volume.
Adding $8$ shells (our default setting) further improves the visual quality, as the finer discretization enhances the capability to model more complex geometry and appearance. 
The performance gain stagnates with even more shells, likely because the model struggles to converge when the feature map becomes too large. 
\Cref{fig:shell_visualization} shows an example identity of what each Gaussian shell has learned:
progressing from the innermost to the outermost shell, every shell captures some texture details, and the outer shell models details that deviate from the base template, such as hat and hair.
\paragraph{Mask and Part Discriminator} The mask discriminator involves the inclusion of the alpha channel as part of the input to the RGB images and is very beneficial for disambiguating background by learning plausible human silhouettes. As demonstrated in \Cref{fig:ablation}, the model trained without mask discriminator generates disturbing white artifacts outside the human shape. 
\Cref{fig:ablation} also shows the improvement in small features in the facial area and better structure for hands and feet introduced by facial and feet/hands discriminator.
We also quantify the joint improvement of the mask and part discriminator and observed an FID boost from 20.63 to 18.7 at \(128^2\) resolution.

\paragraph{Types of Gaussian Anchors.} 
In our representation, we opt to sample the Gaussians on shell meshes. 
There are other alternatives to sample the Gaussians, including constructing tetrahedra akin to the original Shell Map proposed by Porumbescu \etal~\cite{10.1145/1186822.1073239}; or completely discarding the template by sampling uniformly inside a bounding box. We compared with these alternatives and found the shell representation achieves the best visual quality and convergence speed. 
A more detailed discussion can be found in \supp.

%% file: tex/conclusion.tex
\vspace{0.2cm}

\section{Discussion}\label{sec:conclusion}

\paragraph{Limitations.} 
Our method is not without limitations. 
Like almost all existing 3D human generation approaches, our method also relies on a parametric deformation model to enable articulation, which falls short in handling the dynamics of hair and loose clothing. 
Due to the irregularity and sparsity of Gaussians, it is not obvious how to extract the accurate geometry and normals. Surface splatting, with a stronger surface prior may be a potential solution, which we plan to investigate in the future.
Lastly, while ranking among the best methods, our current results still cannot achieve photorealism. The challenge stems from the complexity of human appearance. A promising direction is to combine a small amount of multi-view studio captures with in-the-wild datasets to leverage priors discovered in controlled and richly annotated datasets.
\paragraph{Ethical Concerns.}
GANs, like the one we developed, carry the risk of being utilized for creating altered images of real individuals. This inappropriate application of image generation methods represents a danger to society, and we firmly oppose the use of our research for disseminating false information or damaging reputations. Additionally, we acknowledge that there might be a deficit in diversity in our outcomes, which could be a consequence of inherent biases in the datasets we utilize.
\paragraph{Conclusions.}
In this work, we present Gaussian Shell Maps (\moniker{}s), a novel framework that effectively combines CNN-based generators with 3D Gaussian rendering primitives for the generation of digital humans. 
Our experiments demonstrate the potential of \moniker{}s in generating diverse and detailed human figures, including complex features like clothing and accessories. The framework shows promise in enhancing the efficiency of rendering processes, a crucial factor for real-time applications in industries like virtual reality and cinematic production. 
The potential of \moniker{}s in practical scenarios invites further exploration, and continued research could enhance digital human modeling.

%% file: tex/supplement.tex
\twocolumn[{%
\renewcommand\twocolumn[1][]{#1}%
\maketitle
\begin{center}
\textbf{\Large Supplementary Materials}
\end{center}

\thispagestyle{empty}
\begin{center}
     \includegraphics[width=\linewidth]{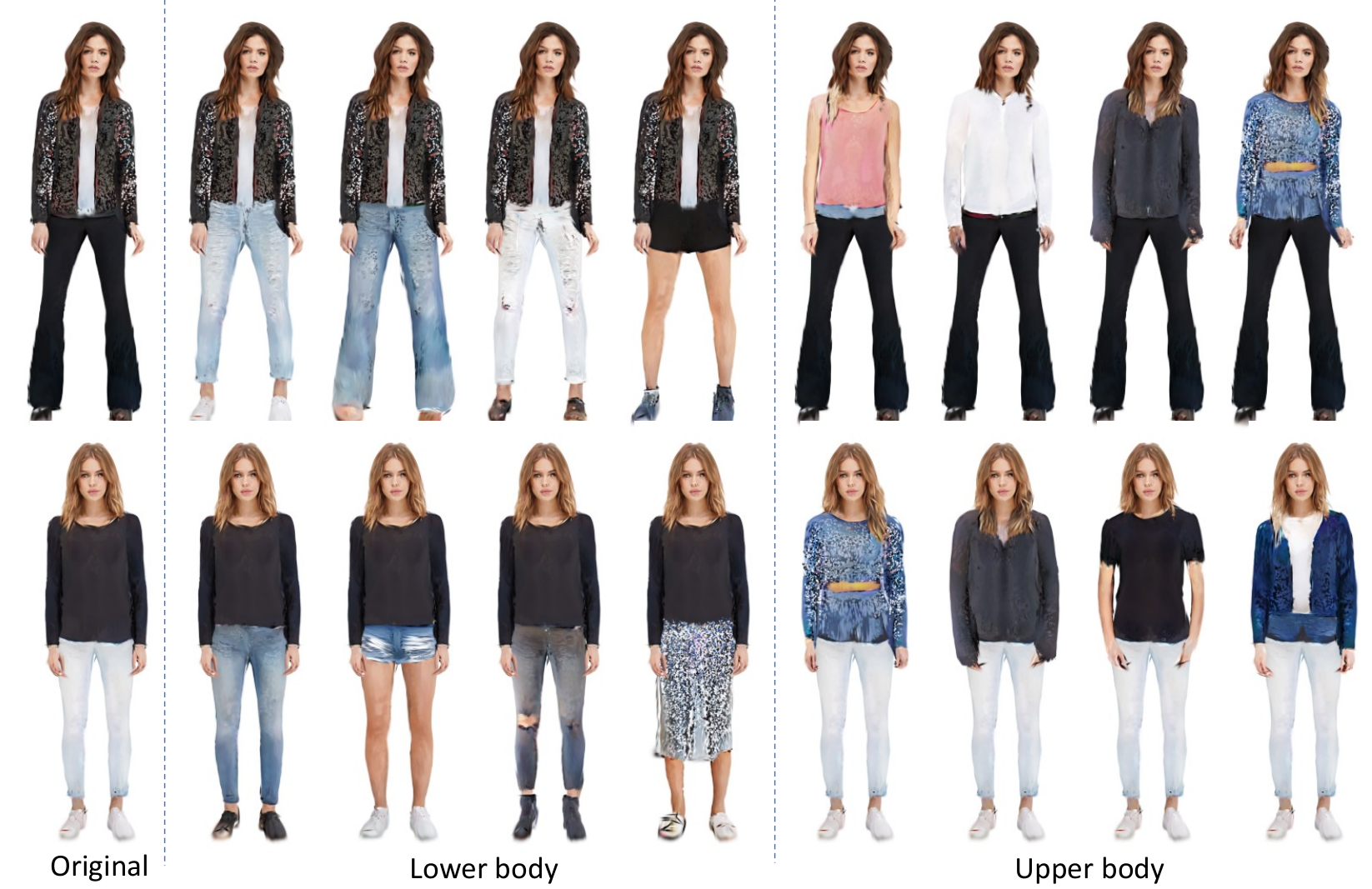}
     \captionof{figure}{\textbf{Appearance Editing.} 3D Gaussians offer an explicit representation, thereby facilitating convenient post-generation editing. In this example, we demonstrate swapping the clothing of two generated identities. Please refer to~\Cref{sec:editing} for further details.}\label{fig:editing}
\end{center}
}]

\begin{figure*}
\centering
\vspace{-0.9cm}
\noindent\includegraphics[width=\linewidth]{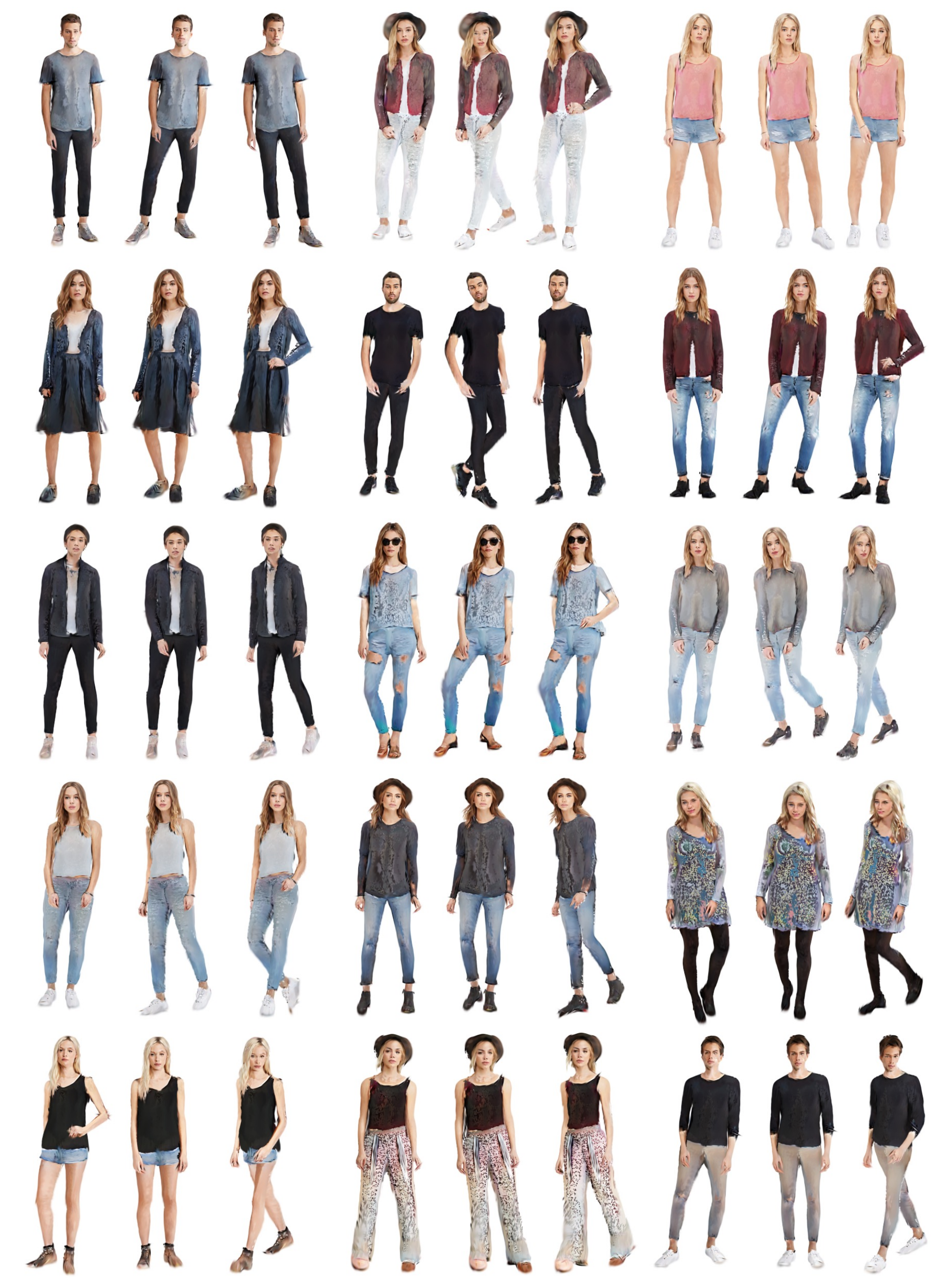}
\caption{\textbf{Visualization.} 3D humans rendered in different poses using our GSM method. }
\label{fig:vis}
\end{figure*}

\begin{figure*}
\centering
\noindent\includegraphics[width=\linewidth]{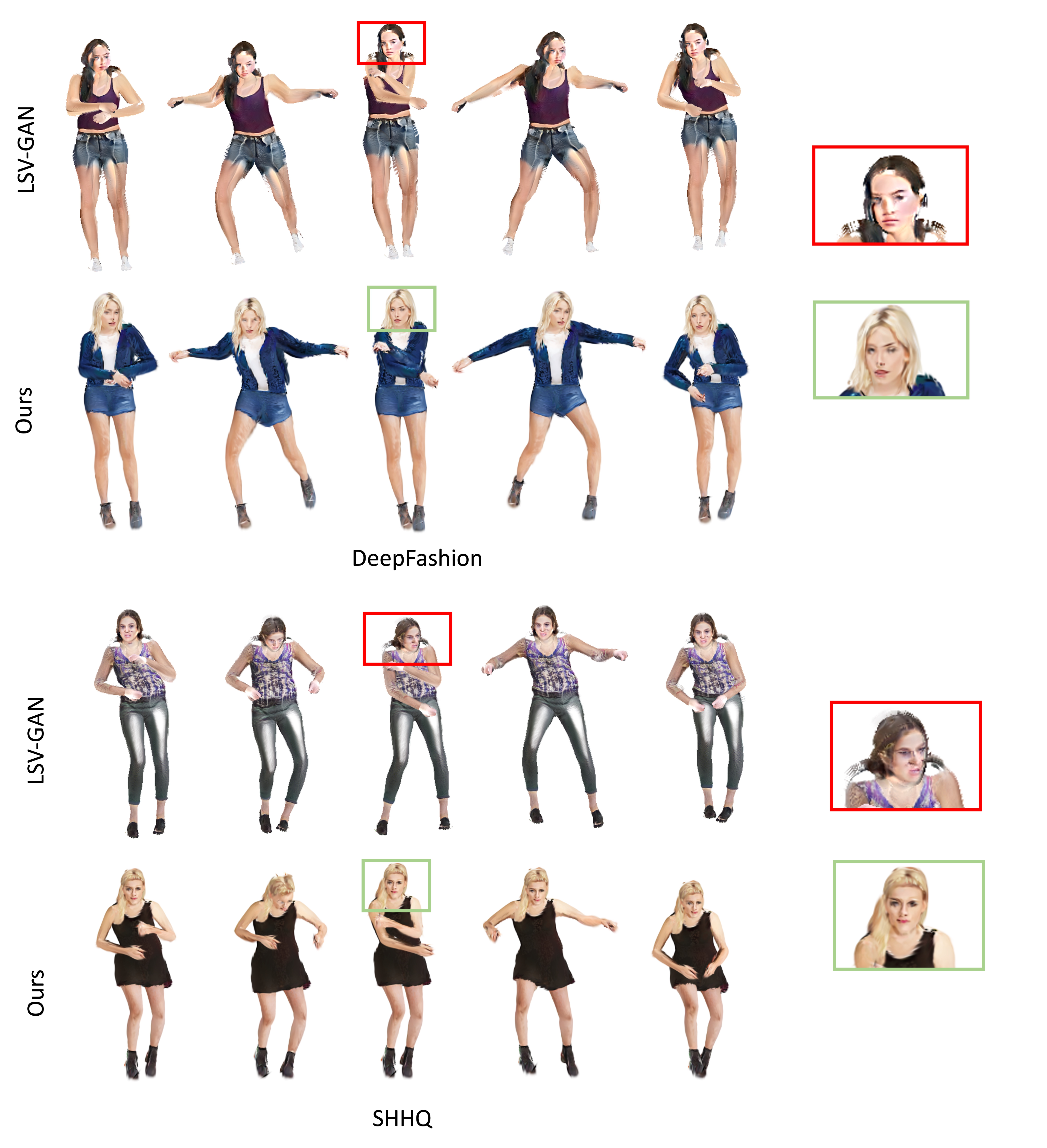}
\caption{\textbf{LSV comparison.}  LSV-GAN~\cite{LayeredSurfaceVolumes} suffers from discontinuities and facial artifacts in DeepFashion and background bleeding into textures in SHHQ. In comparison, our results show high-quality facial details and consistency. }
\label{fig:lsv}
\end{figure*}

\begin{figure*}
\centering
\noindent\includegraphics[width=\linewidth]{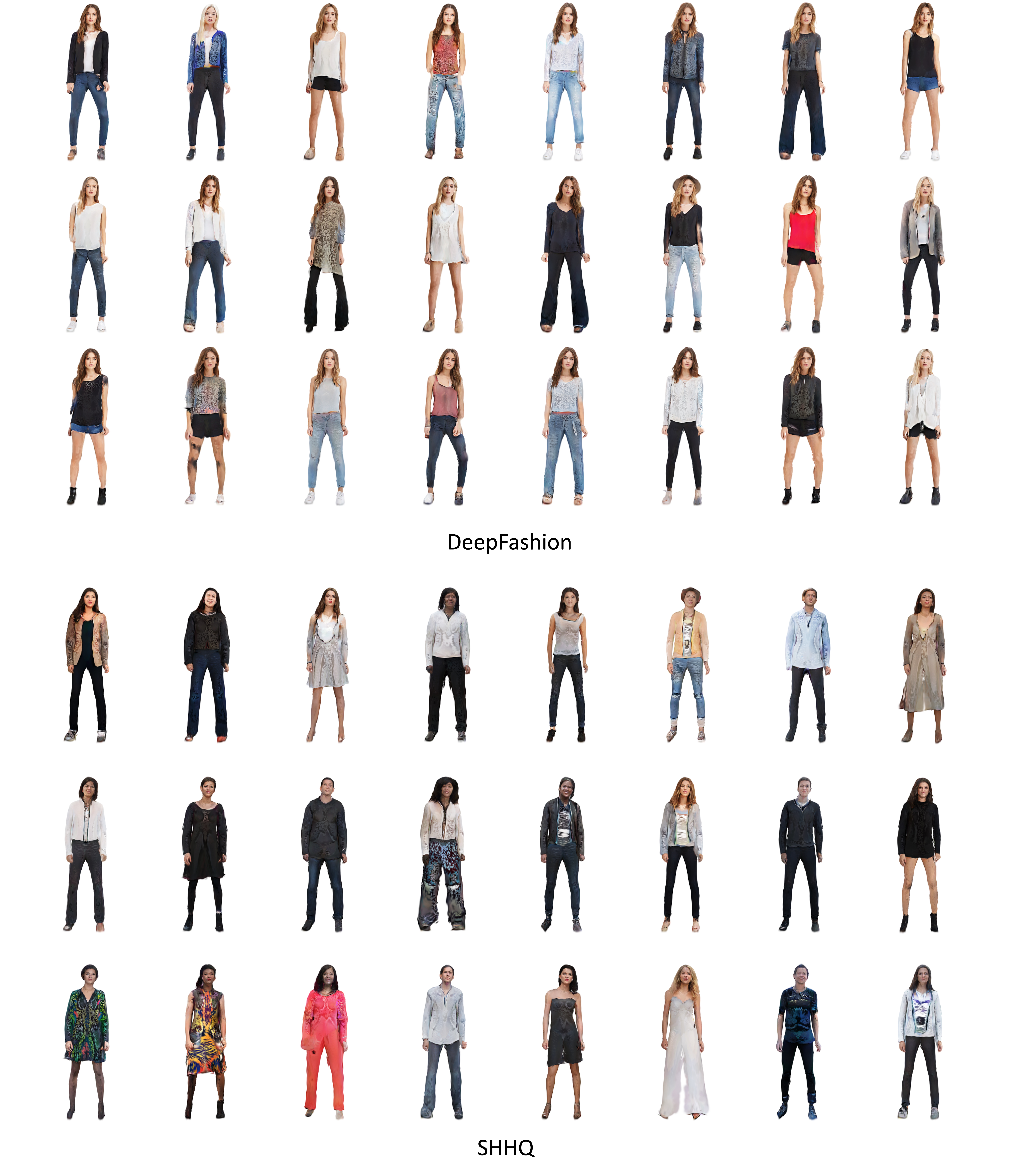}
\caption{\textbf{Random Samples.} Randomly generated samples of 3D humans under same pose using or GSM method trained on DeepFashion~\cite{liu2016deepfasion} and SHHQ~\cite{fu2022stylehuman} datasets.}
\label{fig:uncurated_deepfashion}
\end{figure*}

\section{Further Analysis}
\subsection{Types of Gaussian Anchors.}
\moniker{} anchors the 3D Gaussians on shell meshes. We evaluated multiple alternative ways to anchor the Gaussians, testing the following three methods at \(128^2\) resolution after training with \(1.6k\) Kimgs:
\begin{inparaenum}[(i)]
\item \textbf{in bounding box}: The Gaussians are uniformly sampled within the bounding box of the 3D human mesh, with Gaussian features interpolated from axis-aligned triplane features. This variant differs from our proposed \moniker{} as it does not utilize the shell map to learn features in texture space. Instead, 3D Gaussian features are learned in world space, requiring the generator to also model the distribution of diverse human body poses. With this variant, we demonstrate the importance of using the shell map.
\item \textbf{on a single shell with learned offset}: This variant samples only on the base mesh, the SMPL mesh, but allows deviations from the mesh template by applying a learned offset per Gaussian, predicted by the generator as part of the feature textures. This approach emulates the typical pipeline of existing 3D human GANs, where clothing and hair are captured by offsetting the template unclothed mesh. 
\item \textbf{in tets}: The Gaussians are sampled not only on the shell meshes but also in between them in tetrahedra, constructed by connecting mesh vertices. This variant is more akin to the original Shell Map proposed by Porumbescu \etal. 
\end{inparaenum}
As shown in \Cref{tab:ablation_anchoring}, the bounding box variant underperforms, as the generator struggles to handle deformation jointly with appearance. Learning offsets to model surface details different from the template mesh yields subpar quality. This suggests that varying the Gaussians' positions complicates the already non-convex optimization problem, as the positions are highly correlated with the rest of the Gaussian properties. Finally, sampling in tets shows slow convergence and does not improve the FID. Additionally, this model exhibits a slower rendering speed (speed for deformation and rasterization for a generated identity) of 20 ms/img versus 9 ms/img.

\begin{table}[h]
\caption{
\textbf{Anchoring types.} Ablation on the anchoring type performed on $128^2$ resolution trained for $1.6k$ KImgs on SHHQ.}
\label{tab:ablation_anchoring}
\begin{minipage}{\columnwidth}
\begin{center}
\begin{tabular}{rcccc}
\toprule
Anchoring & bbox & tets & \makecell{learned\\offset} & \makecell{triangles\\(proposed)} \\\midrule
FID $\downarrow$ & 63.90 & 24.66  & 29.30 & \textbf{20.63}  \\
\bottomrule
\end{tabular}
\end{center}
\end{minipage}
\end{table}%

\subsection{Sampling Densities}
We evaluate the effect of the number of Gaussians on the generation quality.
For this study, we train on $512^2$ resolution and evaluate the FID score after training with $10k$ KImgs.
Since the sampling density will affect the Gaussian scale, we adjust the scaling regularization and initialization accordingly.
As shown in~\Cref{tab:ablation_num_gauss}, using $100$K Gaussians yields empirically the best result in terms of FID for the SHHQ dataset.
Using $50k$ Gaussian samples yields the highest FID, suggesting that Gaussians are likely too few to fully model the appearance complexity exhibited in the dataset.
On the other hand, using too many Gaussians, \eg $200k$, can harm the FID. We observe that this drop under a high sampling density scenario is due to the tendency of Gaussians to learn small scales while modeling high-frequency details. This adds complexity to the already challenging task of optimizing opacity and scaling. As a result, we might notice unwanted dotted patterns, especially in cloth areas. The FID score easily detects such an unnatural appearance.  

\begin{table}[h]
\caption{ 
\textbf{Ablation: Number of Gaussians.}  Ablation on the number of Gaussians performed on $512^2$ resolution trained for $10$K KImgs on SHHQ dataset.}
\label{tab:ablation_num_gauss}
\begin{minipage}{\columnwidth}
\begin{center}
\begin{tabular}{rccc}
\toprule
Number & $50k$ & $100k$ & $200k$\\\midrule
FID $\downarrow$ & 23.83 & 13.30  & 19.96  \\
\bottomrule
\end{tabular}
\end{center}
\end{minipage}
\end{table}%

\subsection{Relation and Comparison with LSV-GAN}
Concurrent work, LSV-GAN~\cite{LayeredSurfaceVolumes}, also employs shell meshes and rasterization to efficiently model diverse human shapes and appearances. Our approach, however, distinguishes itself from LSV-GAN by populating the shell meshes with 3D Gaussians and employing differentiable Gaussian Splatting for rendering~\cite{kerbl3Dgaussians}. The spatial span of these Gaussians fills the space between shells with continuous functions.
As illustrated in~\Cref{fig:lsv}, LSV-GAN often exhibits artifacts at silhouette boundaries due to its discontinuous representation of shell volume. In contrast, our method yields smoother and more natural boundaries. Moreover, the utilization of 3D Gaussians allows us to define RGB and alpha values beyond the boundary shell, effectively expanding our capability to model deviations from the template mesh. This leads to more varied geometry and appearances of the human body, including loose clothing and accessories, as seen in Figure 1 of the main manuscript and \Cref{fig:vis} in this document.

\section{Additional Qualitative Results}
In this section, we present further qualitative results. All visual examples have been sampled using the truncation technique detailed in EG3D~\cite{EG3D}. Additional animated results can be found on the \href{https://rameenabdal.github.io/GaussianShellMaps/}{webpage}.

\subsection{Random Samples}
To showcase the quality and diversity of our GSM method, we display randomly sampled results under identical poses in~\Cref{fig:uncurated_deepfashion}. For both the DeepFashion~\cite{liu2016deepfasion} and SHHQ~\cite{fu2022stylehuman} datasets, our method successfully generates a variety of body shapes, accessories like hats, loose clothing, and intricate details on clothes.

\subsection{Articulation and Novel View Rendering}
On the \href{https://rameenabdal.github.io/GaussianShellMaps/}{webpage}, we present videos demonstrating articulation and novel view rendering results. The articulation sequences are provided by the AMASS~\cite{AMASS:ICCV:2019} dataset. 
Notably, our method avoids the temporal flickering artifacts common in other models, as it directly renders at the target resolution. This efficiency is due to our use of rasterization instead of the more costly volumetric rendering approach.

\subsection{Appearance Editing}\label{sec:editing}
A significant advantage of explicit representations like 3D Gaussians, especially when compared to implicit representations such as radiance fields, is their enhanced editability. In~\Cref{fig:editing}, we illustrate this benefit through a redressing application, where we interchange the upper and lower body appearances between multiple generated instances. This editing process involves selecting Gaussians within a specific region (\eg, lower or upper body) and then swapping their properties with those from another instance. This method is feasible because the Gaussian positions are anchored on the shell meshes and remain consistent across instances, with the appearance being defined solely by their properties.

%% file: main.bbl
\begin{thebibliography}{78}
\providecommand{\natexlab}[1]{#1}
\providecommand{\url}[1]{\texttt{#1}}
\expandafter\ifx\csname urlstyle\endcsname\relax
  \providecommand{\doi}[1]{doi: #1}\else
  \providecommand{\doi}{doi: \begingroup \urlstyle{rm}\Url}\fi

\bibitem[Aliev et~al.(2020)Aliev, Sevastopolsky, Kolos, Ulyanov, and Lempitsky]{aliev2020neural}
Kara-Ali Aliev, Artem Sevastopolsky, Maria Kolos, Dmitry Ulyanov, and Victor Lempitsky.
\newblock Neural point-based graphics.
\newblock In \emph{ECCV}, 2020.

\bibitem[An et~al.(2023)An, Xu, Shi, Song, Ogras, and Luo]{an2023panohead}
Sizhe An, Hongyi Xu, Yichun Shi, Guoxian Song, Umit~Y Ogras, and Linjie Luo.
\newblock Panohead: Geometry-aware 3d full-head synthesis in 360deg.
\newblock In \emph{CVPR}, 2023.

\bibitem[Aneja et~al.(2022)Aneja, Thies, Dai, and Nießner]{aneja2022clipface}
Shivangi Aneja, Justus Thies, Angela Dai, and Matthias Nießner.
\newblock {C}lip{F}ace: {T}ext-guided {E}diting of {T}extured 3{D} {M}orphable {M}odels.
\newblock In \emph{ArXiv preprint arXiv:2212.01406}, 2022.

\bibitem[Balaji et~al.(2022)Balaji, Nah, Huang, Vahdat, Song, Zhang, Kreis, Aittala, Aila, Laine, Catanzaro, Karras, and Liu]{Balaji2022eDiffITD}
Yogesh Balaji, Seungjun Nah, Xun Huang, Arash Vahdat, Jiaming Song, Qinsheng Zhang, Karsten Kreis, Miika Aittala, Timo Aila, Samuli Laine, Bryan Catanzaro, Tero Karras, and Ming-Yu Liu.
\newblock ediff-i: Text-to-image diffusion models with an ensemble of expert denoisers.
\newblock \emph{ArXiv}, abs/2211.01324, 2022.

\bibitem[Barron et~al.(2021)Barron, Mildenhall, Tancik, Hedman, Martin-Brualla, and Srinivasan]{mip-NeRF}
Jonathan~T Barron, Ben Mildenhall, Matthew Tancik, Peter Hedman, Ricardo Martin-Brualla, and Pratul~P Srinivasan.
\newblock Mip-nerf: A multiscale representation for anti-aliasing neural radiance fields.
\newblock In \emph{ICCV}, 2021.

\bibitem[Bergman et~al.(2022)Bergman, Kellnhofer, Yifan, Chan, Lindell, and Wetzstein]{bergman2022generative}
Alexander Bergman, Petr Kellnhofer, Wang Yifan, Eric Chan, David Lindell, and Gordon Wetzstein.
\newblock Generative neural articulated radiance fields.
\newblock \emph{NeurIPS}, 2022.

\bibitem[Bergman et~al.(2023)Bergman, Yifan, and Wetzstein]{Bergman2023Articulated3H}
Alexander~W. Bergman, Wang Yifan, and Gordon Wetzstein.
\newblock Articulated 3d head avatar generation using text-to-image diffusion models.
\newblock 2023.

\bibitem[Cao et~al.(2023)Cao, Cao, Han, Shan, and Wong]{cao2023dreamavatar}
Yukang Cao, Yan-Pei Cao, Kai Han, Ying Shan, and Kwan-Yee~K. Wong.
\newblock Dreamavatar: Text-and-shape guided 3d human avatar generation via diffusion models, 2023.

\bibitem[Chan et~al.(2023)Chan, Nagano, Chan, Bergman, Park, Levy, Aittala, Mello, Karras, and Wetzstein]{Chan2023GenerativeNV}
Eric Chan, Koki Nagano, Matthew~A. Chan, Alexander~W. Bergman, Jeong~Joon Park, Axel Levy, Miika Aittala, Shalini~De Mello, Tero Karras, and Gordon Wetzstein.
\newblock Generative novel view synthesis with 3d-aware diffusion models.
\newblock \emph{ArXiv}, abs/2304.02602, 2023.

\bibitem[Chan et~al.(2021)Chan, Monteiro, Kellnhofer, Wu, and Wetzstein]{piGAN}
Eric~R Chan, Marco Monteiro, Petr Kellnhofer, Jiajun Wu, and Gordon Wetzstein.
\newblock pi-gan: Periodic implicit generative adversarial networks for 3d-aware image synthesis.
\newblock In \emph{CVPR}, 2021.

\bibitem[Chan et~al.(2022)Chan, Lin, Chan, Nagano, Pan, Mello, Gallo, Guibas, Tremblay, Khamis, Karras, and Wetzstein]{EG3D}
Eric~R. Chan, Connor~Z. Lin, Matthew~A. Chan, Koki Nagano, Boxiao Pan, Shalini~De Mello, Orazio Gallo, Leonidas Guibas, Jonathan Tremblay, Sameh Khamis, Tero Karras, and Gordon Wetzstein.
\newblock Efficient geometry-aware {3D} generative adversarial networks.
\newblock In \emph{CVPR}, 2022.

\bibitem[Chen et~al.(2023{\natexlab{a}})Chen, Chen, Jiao, and Jia]{Chen2023Fantasia3DDG}
Rui Chen, Y. Chen, Ningxin Jiao, and Kui Jia.
\newblock Fantasia3d: Disentangling geometry and appearance for high-quality text-to-3d content creation.
\newblock \emph{ArXiv}, abs/2303.13873, 2023{\natexlab{a}}.

\bibitem[Chen et~al.(2023{\natexlab{b}})Chen, Huang, Bin, Yu, and Liao]{chen2023veri3d}
Xinya Chen, Jiaxin Huang, Yanrui Bin, Lu Yu, and Yiyi Liao.
\newblock Veri3d: Generative vertex-based radiance fields for 3d controllable human image synthesis, 2023{\natexlab{b}}.

\bibitem[Diolatzis et~al.(2023)Diolatzis, Novak, Rousselle, Granskog, Aittala, Ramamoorthi, and Drettakis]{DNRGARD23}
Stavros Diolatzis, Jan Novak, Fabrice Rousselle, Jonathan Granskog, Miika Aittala, Ravi Ramamoorthi, and George Drettakis.
\newblock Mesogan: Generative neural reflectance shells.
\newblock \emph{Comput. Graph. Forum}, 2023.

\bibitem[Dong et~al.(2023)Dong, Chen, Yang, J.Black, Hilliges, and Geiger]{dong2023ag3d}
Zijian Dong, Xu Chen, Jinlong Yang, Michael J.Black, Otmar Hilliges, and Andreas Geiger.
\newblock {AG3D}: Learning to generate {3D} avatars from {2D} image collections.
\newblock In \emph{ICCV}, 2023.

\bibitem[Fu et~al.(2022{\natexlab{a}})Fu, Li, Jiang, Lin, Qian, Loy, Wu, and Liu]{fu2022stylegan}
Jianglin Fu, Shikai Li, Yuming Jiang, Kwan-Yee Lin, Chen Qian, Chen~Change Loy, Wayne Wu, and Ziwei Liu.
\newblock Stylegan-human: A data-centric odyssey of human generation.
\newblock In \emph{ECCV}, 2022{\natexlab{a}}.

\bibitem[Fu et~al.(2022{\natexlab{b}})Fu, Li, Jiang, Lin, Qian, Loy, Wu, and Liu]{fu2022stylehuman}
Jianglin Fu, Shikai Li, Yuming Jiang, Kwan-Yee Lin, Chen Qian, Chen~Change Loy, Wayne Wu, and Ziwei Liu.
\newblock Stylegan-human: A data-centric odyssey of human generation.
\newblock In \emph{ECCV}, 2022{\natexlab{b}}.

\bibitem[Gao et~al.(2022)Gao, Shen, Wang, Chen, Yin, Li, Litany, Gojcic, and Fidler]{gao2022get3d}
Jun Gao, Tianchang Shen, Zian Wang, Wenzheng Chen, Kangxue Yin, Daiqing Li, Or Litany, Zan Gojcic, and Sanja Fidler.
\newblock Get3d: A generative model of high quality 3d textured shapes learned from images.
\newblock \emph{NeurIPS}, 2022.

\bibitem[Grigorev et~al.(2021)Grigorev, Iskakov, Ianina, Bashirov, Zakharkin, Vakhitov, and Lempitsky]{grigorev2021stylepeople}
Artur Grigorev, Karim Iskakov, Anastasia Ianina, Renat Bashirov, Ilya Zakharkin, Alexander Vakhitov, and Victor Lempitsky.
\newblock Stylepeople: A generative model of fullbody human avatars.
\newblock In \emph{CVPR}, 2021.

\bibitem[Gross and Pfister(2011)]{gross2011point}
Markus Gross and Hanspeter Pfister.
\newblock \emph{Point-based graphics}.
\newblock Elsevier, 2011.

\bibitem[Grossman and Dally(1998)]{grossman1998point}
Jeffrey~P Grossman and William~J Dally.
\newblock Point sample rendering.
\newblock In \emph{Proceedings of the Eurographics Workshop}, 1998.

\bibitem[Gu et~al.(2022)Gu, Liu, Wang, and Theobalt]{StyleNeRF}
Jiatao Gu, Lingjie Liu, Peng Wang, and Christian Theobalt.
\newblock Stylenerf: A style-based 3d aware generator for high-resolution image synthesis.
\newblock In \emph{ICLR}, 2022.

\bibitem[Hong et~al.(2023)Hong, Chen, LAN, Pan, and Liu]{hong2023evad}
Fangzhou Hong, Zhaoxi Chen, Yushi LAN, Liang Pan, and Ziwei Liu.
\newblock {EVA}3d: Compositional 3d human generation from 2d image collections.
\newblock In \emph{ICLR}, 2023.

\bibitem[Jiang et~al.(2023)Jiang, Jiang, Wang, Luo, Chen, and Xu]{jiang2023humangen}
Suyi Jiang, Haoran Jiang, Ziyu Wang, Haimin Luo, Wenzheng Chen, and Lan Xu.
\newblock Humangen: Generating human radiance fields with explicit priors.
\newblock In \emph{CVPR}, 2023.

\bibitem[Karras et~al.(2019)Karras, Laine, and Aila]{StyleGAN}
Tero Karras, Samuli Laine, and Timo Aila.
\newblock A style-based generator architecture for generative adversarial networks.
\newblock In \emph{CVPR}, 2019.

\bibitem[Karras et~al.(2020)Karras, Laine, Aittala, Hellsten, Lehtinen, and Aila]{Karras2019stylegan2}
Tero Karras, Samuli Laine, Miika Aittala, Janne Hellsten, Jaakko Lehtinen, and Timo Aila.
\newblock Analyzing and improving the image quality of {StyleGAN}.
\newblock In \emph{Proc. CVPR}, 2020.

\bibitem[Karras et~al.(2021)Karras, Aittala, Laine, H\"ark\"onen, Hellsten, Lehtinen, and Aila]{stylegan3}
Tero Karras, Miika Aittala, Samuli Laine, Erik H\"ark\"onen, Janne Hellsten, Jaakko Lehtinen, and Timo Aila.
\newblock Alias-free generative adversarial networks.
\newblock In \emph{NeurIPS}, 2021.

\bibitem[Kerbl et~al.(2023)Kerbl, Kopanas, Leimk{\"u}hler, and Drettakis]{kerbl3Dgaussians}
Bernhard Kerbl, Georgios Kopanas, Thomas Leimk{\"u}hler, and George Drettakis.
\newblock 3d gaussian splatting for real-time radiance field rendering.
\newblock \emph{ACM TOG}, 2023.

\bibitem[Kolotouros et~al.(2023)Kolotouros, Alldieck, Zanfir, Bazavan, Fieraru, and Sminchisescu]{Kolotouros2023DreamHumanA3}
Nikos Kolotouros, Thiemo Alldieck, Andrei Zanfir, Eduard~Gabriel Bazavan, Mihai Fieraru, and Cristian Sminchisescu.
\newblock Dreamhuman: Animatable 3d avatars from text.
\newblock \emph{ArXiv}, abs/2306.09329, 2023.

\bibitem[Kopanas et~al.(2021)Kopanas, Philip, Leimk{\"u}hler, and Drettakis]{kopanas2021point}
Georgios Kopanas, Julien Philip, Thomas Leimk{\"u}hler, and George Drettakis.
\newblock Point-based neural rendering with per-view optimization.
\newblock In \emph{Comput. Graph. Forum}, 2021.

\bibitem[Laine and Karras(2011)]{laine2011high}
Samuli Laine and Tero Karras.
\newblock High-performance software rasterization on gpus.
\newblock In \emph{Proceedings of the ACM SIGGRAPH Symposium on High Performance Graphics}, 2011.

\bibitem[Li et~al.(2021)Li, Yang, Ross, and Kanazawa]{li2021aist}
Ruilong Li, Shan Yang, David~A. Ross, and Angjoo Kanazawa.
\newblock Learn to dance with aist++: Music conditioned 3d dance generation, 2021.

\bibitem[Lin et~al.(2022)Lin, Gao, Tang, Takikawa, Zeng, Huang, Kreis, Fidler, Liu, and Lin]{Lin2022Magic3DHT}
Chen-Hsuan Lin, Jun Gao, Luming Tang, Towaki Takikawa, Xiaohui Zeng, Xun Huang, Karsten Kreis, Sanja Fidler, Ming-Yu Liu, and Tsung-Yi Lin.
\newblock Magic3d: High-resolution text-to-3d content creation.
\newblock \emph{ArXiv}, abs/2211.10440, 2022.

\bibitem[Liu et~al.(2023)Liu, Wu, Hoorick, Tokmakov, Zakharov, and Vondrick]{Liu2023Zero1to3ZO}
Ruoshi Liu, Rundi Wu, Basile~Van Hoorick, Pavel Tokmakov, Sergey Zakharov, and Carl Vondrick.
\newblock Zero-1-to-3: Zero-shot one image to 3d object.
\newblock \emph{ArXiv}, abs/2303.11328, 2023.

\bibitem[Liu et~al.(2016)Liu, Luo, Qiu, Wang, and Tang]{liu2016deepfasion}
Ziwei Liu, Ping Luo, Shi Qiu, Xiaogang Wang, and Xiaoou Tang.
\newblock Deepfashion: Powering robust clothes recognition and retrieval with rich annotations.
\newblock In \emph{CVPR}, 2016.

\bibitem[Loper et~al.(2015)Loper, Mahmood, Romero, Pons-Moll, and Black]{loper2015smpl}
Matthew Loper, Naureen Mahmood, Javier Romero, Gerard Pons-Moll, and Michael~J Black.
\newblock Smpl: A skinned multi-person linear model.
\newblock \emph{ACM TOG}, 2015.

\bibitem[Mahmood et~al.(2019)Mahmood, Ghorbani, Troje, Pons-Moll, and Black]{AMASS:ICCV:2019}
Naureen Mahmood, Nima Ghorbani, Nikolaus~F. Troje, Gerard Pons-Moll, and Michael~J. Black.
\newblock {AMASS}: Archive of motion capture as surface shapes.
\newblock In \emph{International Conference on Computer Vision}, pages 5442--5451, 2019.

\bibitem[Mejjati et~al.(2021)Mejjati, Milefchik, Gokaslan, Wang, Kim, and Tompkin]{GaussiGAN}
Youssef~A Mejjati, Isa Milefchik, Aaron Gokaslan, Oliver Wang, Kwang~In Kim, and James Tompkin.
\newblock Gaussigan: Controllable image synthesis with 3d gaussians from unposed silhouettes.
\newblock \emph{arXiv preprint arXiv:2106.13215}, 2021.

\bibitem[Niemeyer and Geiger(2021{\natexlab{a}})]{CAMPARI}
Michael Niemeyer and Andreas Geiger.
\newblock Campari: Camera-aware decomposed generative neural radiance fields.
\newblock In \emph{3DV}, 2021{\natexlab{a}}.

\bibitem[Niemeyer and Geiger(2021{\natexlab{b}})]{Giraffe}
Michael Niemeyer and Andreas Geiger.
\newblock Giraffe: Representing scenes as compositional generative neural feature fields.
\newblock In \emph{CVPR}, 2021{\natexlab{b}}.

\bibitem[Noguchi et~al.(2022)Noguchi, Sun, Lin, and Harada]{noguchi2022unsupervised}
Atsuhiro Noguchi, Xiao Sun, Stephen Lin, and Tatsuya Harada.
\newblock Unsupervised learning of efficient geometry-aware neural articulated representations.
\newblock In \emph{ECCV}, 2022.

\bibitem[Or-El et~al.(2021)Or-El, Luo, Shan, Shechtman, Park, and Kemelmacher-Shlizerman]{StyleSDF}
Roy Or-El, Xuan Luo, Mengyi Shan, Eli Shechtman, Jeong~Joon Park, and Ira Kemelmacher-Shlizerman.
\newblock Style{SDF}: {H}igh-{R}esolution {3D}-{C}onsistent {I}mage and {G}eometry {G}eneration.
\newblock \emph{arXiv preprint arXiv:2112.11427}, 2021.

\bibitem[Pavlakos et~al.(2019)Pavlakos, Choutas, Ghorbani, Bolkart, Osman, Tzionas, and Black]{SMPL-X:2019}
Georgios Pavlakos, Vasileios Choutas, Nima Ghorbani, Timo Bolkart, Ahmed A.~A. Osman, Dimitrios Tzionas, and Michael~J. Black.
\newblock Expressive body capture: {3D} hands, face, and body from a single image.
\newblock In \emph{CVPR}, 2019.

\bibitem[Po and Wetzstein(2023)]{po2023compositional}
Ryan Po and Gordon Wetzstein.
\newblock Compositional 3d scene generation using locally conditioned diffusion.
\newblock \emph{arXiv preprint arXiv:2303.12218}, 2023.

\bibitem[Po et~al.(2023)Po, Yifan, Golyanik, Aberman, Barron, Bermano, Chan, Dekel, Holynski, Kanazawa, Liu, Liu, Mildenhall, Nießner, Ommer, Theobalt, Wonka, and Wetzstein]{po2023state}
Ryan Po, Wang Yifan, Vladislav Golyanik, Kfir Aberman, Jonathan~T. Barron, Amit~H. Bermano, Eric~Ryan Chan, Tali Dekel, Aleksander Holynski, Angjoo Kanazawa, C.~Karen Liu, Lingjie Liu, Ben Mildenhall, Matthias Nießner, Björn Ommer, Christian Theobalt, Peter Wonka, and Gordon Wetzstein.
\newblock State of the art on diffusion models for visual computing, 2023.

\bibitem[Poole et~al.(2022)Poole, Jain, Barron, and Mildenhall]{poole2022dreamfusion}
Ben Poole, Ajay Jain, Jonathan~T Barron, and Ben Mildenhall.
\newblock Dreamfusion: Text-to-3d using 2d diffusion.
\newblock \emph{arXiv preprint arXiv:2209.14988}, 2022.

\bibitem[Porumbescu et~al.(2005)Porumbescu, Budge, Feng, and Joy]{10.1145/1186822.1073239}
Serban~D. Porumbescu, Brian Budge, Louis Feng, and Kenneth~I. Joy.
\newblock Shell maps.
\newblock In \emph{ACM SIGGRAPH}, 2005.

\bibitem[Ramesh et~al.(2022)Ramesh, Dhariwal, Nichol, Chu, and Chen]{DALLE-2}
Aditya Ramesh, Prafulla Dhariwal, Alex Nichol, Casey Chu, and Mark Chen.
\newblock Hierarchical text-conditional image generation with clip latents.
\newblock \emph{arXiv preprint arXiv:2204.06125}, 2022.

\bibitem[Ren et~al.(2002)Ren, Pfister, and Zwicker]{ren2002object}
Liu Ren, Hanspeter Pfister, and Matthias Zwicker.
\newblock Object space ewa surface splatting: A hardware accelerated approach to high quality point rendering.
\newblock In \emph{Comput. Graph. Forum}, 2002.

\bibitem[Rhodin et~al.(2015)Rhodin, Robertini, Richardt, Seidel, and Theobalt]{rhodin2015versatile}
Helge Rhodin, Nadia Robertini, Christian Richardt, Hans-Peter Seidel, and Christian Theobalt.
\newblock A versatile scene model with differentiable visibility applied to generative pose estimation.
\newblock In \emph{ICCV}, 2015.

\bibitem[Rombach et~al.(2021)Rombach, Blattmann, Lorenz, Esser, and Ommer]{Rombach2021HighResolutionIS}
Robin Rombach, A. Blattmann, Dominik Lorenz, Patrick Esser, and Bj{\"o}rn Ommer.
\newblock High-resolution image synthesis with latent diffusion models.
\newblock \emph{CVPR}, 2021.

\bibitem[R{\"u}ckert et~al.(2022)R{\"u}ckert, Franke, and Stamminger]{ruckert2022adop}
Darius R{\"u}ckert, Linus Franke, and Marc Stamminger.
\newblock Adop: Approximate differentiable one-pixel point rendering.
\newblock \emph{ACM TOG}, 2022.

\bibitem[Saharia et~al.(2022)Saharia, Chan, Saxena, Li, Whang, Denton, Ghasemipour, Ayan, Mahdavi, Lopes, Salimans, Ho, Fleet, and Norouzi]{Saharia2022PhotorealisticTD}
Chitwan Saharia, William Chan, Saurabh Saxena, Lala Li, Jay Whang, Emily~L. Denton, Seyed Kamyar~Seyed Ghasemipour, Burcu~Karagol Ayan, Seyedeh~Sara Mahdavi, Raphael~Gontijo Lopes, Tim Salimans, Jonathan Ho, David~J. Fleet, and Mohammad Norouzi.
\newblock Photorealistic text-to-image diffusion models with deep language understanding.
\newblock \emph{ArXiv}, abs/2205.11487, 2022.

\bibitem[Sainz and Pajarola(2004)]{sainz2004point}
Miguel Sainz and Renato Pajarola.
\newblock Point-based rendering techniques.
\newblock \emph{Computers \& Graphics}, 28\penalty0 (6):\penalty0 869--879, 2004.

\bibitem[Sandler et~al.(2018)Sandler, Howard, Zhu, Zhmoginov, and Chen]{sandler2018mobilenetv2}
Mark Sandler, Andrew Howard, Menglong Zhu, Andrey Zhmoginov, and Liang-Chieh Chen.
\newblock Mobilenetv2: Inverted residuals and linear bottlenecks.
\newblock In \emph{CVPR}, 2018.

\bibitem[Sch{\"u}tz et~al.(2022)Sch{\"u}tz, Kerbl, and Wimmer]{schutz2022software}
Markus Sch{\"u}tz, Bernhard Kerbl, and Michael Wimmer.
\newblock Software rasterization of 2 billion points in real time.
\newblock \emph{Proceedings of the ACM on Computer Graphics and Interactive Techniques}, 2022.

\bibitem[Schwarz et~al.(2020)Schwarz, Liao, Niemeyer, and Geiger]{GRAF}
Katja Schwarz, Yiyi Liao, Michael Niemeyer, and Andreas Geiger.
\newblock Graf: Generative radiance fields for 3d-aware image synthesis.
\newblock In \emph{NeurIPS}, 2020.

\bibitem[Sin et~al.(2023)Sin, Ng, and Leong]{sin2023nerfahedron}
Zackary P.~T. Sin, Peter H.~F. Ng, and Hong~Va Leong.
\newblock Nerfahedron: A primitive for animatable neural rendering with interactive speed.
\newblock \emph{Proceedings of the ACM on Computer Graphics and Interactive Techniques}, 2023.

\bibitem[Skorokhodov et~al.(2023)Skorokhodov, Siarohin, Xu, Ren, Lee, Wonka, and Tulyakov]{skorokhodov20233d}
Ivan Skorokhodov, Aliaksandr Siarohin, Yinghao Xu, Jian Ren, Hsin-Ying Lee, Peter Wonka, and Sergey Tulyakov.
\newblock 3d generation on imagenet.
\newblock In \emph{ICLR}, 2023.

\bibitem[Sorkine et~al.(2004)Sorkine, Cohen-Or, Lipman, Alexa, R{\"o}ssl, and Seidel]{sorkine2004laplacian}
Olga Sorkine, Daniel Cohen-Or, Yaron Lipman, Marc Alexa, Christian R{\"o}ssl, and H-P Seidel.
\newblock Laplacian surface editing.
\newblock In \emph{Eurographics}, 2004.

\bibitem[Stoll et~al.(2011)Stoll, Hasler, Gall, Seidel, and Theobalt]{stoll2011fast}
Carsten Stoll, Nils Hasler, Juergen Gall, Hans-Peter Seidel, and Christian Theobalt.
\newblock Fast articulated motion tracking using a sums of gaussians body model.
\newblock In \emph{ICCV}, 2011.

\bibitem[Sun et~al.(2022)Sun, Wang, Shi, Wang, Wang, and Liu]{ide-3d}
Jingxiang Sun, Xuan Wang, Yichun Shi, Lizhen Wang, Jue Wang, and Yebin Liu.
\newblock Ide-3d: Interactive disentangled editing for high-resolution 3d-aware portrait synthesis.
\newblock \emph{ACM TOG}, 2022.

\bibitem[Sun et~al.(2023)Sun, Wang, Wang, Li, Zhang, Zhang, and Liu]{sun2023next3d}
Jingxiang Sun, Xuan Wang, Lizhen Wang, Xiaoyu Li, Yong Zhang, Hongwen Zhang, and Yebin Liu.
\newblock Next3d: Generative neural texture rasterization for 3d-aware head avatars.
\newblock In \emph{CVPR}, 2023.

\bibitem[Tewari et~al.(2020)Tewari, Fried, Thies, Sitzmann, Lombardi, Sunkavalli, Martin-Brualla, Simon, Saragih, Nie{\ss}ner, et~al.]{tewari2020state}
Ayush Tewari, Ohad Fried, Justus Thies, Vincent Sitzmann, Stephen Lombardi, Kalyan Sunkavalli, Ricardo Martin-Brualla, Tomas Simon, Jason Saragih, Matthias Nie{\ss}ner, et~al.
\newblock State of the art on neural rendering.
\newblock In \emph{Comput. Graph. Forum}, 2020.

\bibitem[Tewari et~al.(2022)Tewari, Thies, Mildenhall, Srinivasan, Tretschk, Yifan, Lassner, Sitzmann, Martin-Brualla, Lombardi, et~al.]{tewari2022advances}
Ayush Tewari, Justus Thies, Ben Mildenhall, Pratul Srinivasan, Edgar Tretschk, Wang Yifan, Christoph Lassner, Vincent Sitzmann, Ricardo Martin-Brualla, Stephen Lombardi, et~al.
\newblock Advances in neural rendering.
\newblock In \emph{Comput. Graph. Forum}, 2022.

\bibitem[Tsalicoglou et~al.(2023)Tsalicoglou, Manhardt, Tonioni, Niemeyer, and Tombari]{Tsalicoglou2023TextMeshGO}
Christina Tsalicoglou, Fabian Manhardt, Alessio Tonioni, Michael Niemeyer, and Federico Tombari.
\newblock Textmesh: Generation of realistic 3d meshes from text prompts.
\newblock \emph{ArXiv}, abs/2304.12439, 2023.

\bibitem[Wiles et~al.(2020)Wiles, Gkioxari, Szeliski, and Johnson]{wiles2020synsin}
Olivia Wiles, Georgia Gkioxari, Richard Szeliski, and Justin Johnson.
\newblock Synsin: End-to-end view synthesis from a single image.
\newblock In \emph{CVPR}, 2020.

\bibitem[Xu et~al.(2022)Xu, Xu, Philip, Bi, Shu, Sunkavalli, and Neumann]{xu2022point}
Qiangeng Xu, Zexiang Xu, Julien Philip, Sai Bi, Zhixin Shu, Kalyan Sunkavalli, and Ulrich Neumann.
\newblock Point-nerf: Point-based neural radiance fields.
\newblock In \emph{CVPR}, 2022.

\bibitem[Xu et~al.(2021)Xu, Peng, Yang, Shen, and Zhou]{VolumeGAN}
Yinghao Xu, Sida Peng, Ceyuan Yang, Yujun Shen, and Bolei Zhou.
\newblock 3d-aware image synthesis via learning structural and textural representations.
\newblock \emph{arXiv preprint arXiv:2112.10759}, 2021.

\bibitem[Xu et~al.(2023)Xu, Chai, Shi, Peng, Skorokhodov, Siarohin, Yang, Shen, Lee, Zhou, et~al.]{xu2022discoscene}
Yinghao Xu, Menglei Chai, Zifan Shi, Sida Peng, Ivan Skorokhodov, Aliaksandr Siarohin, Ceyuan Yang, Yujun Shen, Hsin-Ying Lee, Bolei Zhou, et~al.
\newblock Discoscene: Spatially disentangled generative radiance fields for controllable 3d-aware scene synthesis.
\newblock In \emph{CVPR}, 2023.

\bibitem[Xu et~al.(2024)Xu, Yifan, Bergman, Chai, Zhou, and Wetzstein]{LayeredSurfaceVolumes}
Yinghao Xu, Wang Yifan, Alexander~W. Bergman, Menglei Chai, Bolei Zhou, and Gordon Wetzstein.
\newblock Efficient 3d articulated human generation with layered surface volumes.
\newblock In \emph{3DV}, 2024.

\bibitem[Yang et~al.(2023)Yang, Li, Wu, and Dai]{yang20233dhumangan}
Zhuoqian Yang, Shikai Li, Wayne Wu, and Bo Dai.
\newblock 3dhumangan: 3d-aware human image generation with 3d pose mapping.
\newblock In \emph{ICCV}, 2023.

\bibitem[Yifan et~al.(2019)Yifan, Serena, Wu, {\"O}ztireli, and Sorkine-Hornung]{yifan2019differentiable}
Wang Yifan, Felice Serena, Shihao Wu, Cengiz {\"O}ztireli, and Olga Sorkine-Hornung.
\newblock Differentiable surface splatting for point-based geometry processing.
\newblock \emph{ACM TOG}, 2019.

\bibitem[Zhang et~al.(2023{\natexlab{a}})Zhang, Jiang, Yang, Xu, Shi, Song, Xu, Wang, and Feng]{Avatargen2023}
Jianfeng Zhang, Zihang Jiang, Dingdong Yang, Hongyi Xu, Yichun Shi, Guoxian Song, Zhongcong Xu, Xinchao Wang, and Jiashi Feng.
\newblock Avatargen: A 3d generative model for animatable human avatars.
\newblock \emph{ArXiv}, 2023{\natexlab{a}}.

\bibitem[Zhang et~al.(2022)Zhang, Baek, Rusinkiewicz, and Heide]{zhang2022differentiable}
Qiang Zhang, Seung-Hwan Baek, Szymon Rusinkiewicz, and Felix Heide.
\newblock Differentiable point-based radiance fields for efficient view synthesis.
\newblock In \emph{SIGGRAPH Asia}, 2022.

\bibitem[Zhang et~al.(2023{\natexlab{b}})Zhang, Zhang, Rohan, Xu, Song, Yang, and Feng]{zhang2023getavatar}
Xuanmeng Zhang, Jianfeng Zhang, Chacko Rohan, Hongyi Xu, Guoxian Song, Yi Yang, and Jiashi Feng.
\newblock Getavatar: Generative textured meshes for animatable human avatars.
\newblock In \emph{ICCV}, 2023{\natexlab{b}}.

\bibitem[Zheng et~al.(2023)Zheng, Yifan, Wetzstein, Black, and Hilliges]{zheng2023pointavatar}
Yufeng Zheng, Wang Yifan, Gordon Wetzstein, Michael~J Black, and Otmar Hilliges.
\newblock Pointavatar: Deformable point-based head avatars from videos.
\newblock In \emph{CVPR}, 2023.

\bibitem[Zwicker et~al.(2001)Zwicker, Pfister, Van~Baar, and Gross]{zwicker2001ewa}
Matthias Zwicker, Hanspeter Pfister, Jeroen Van~Baar, and Markus Gross.
\newblock Ewa volume splatting.
\newblock In \emph{Proceedings Visualization}, 2001.

\end{thebibliography}
